\documentclass[]{opendatalab}
\usepackage{xcolor}
\usepackage{longtable}
\usepackage{makecell}
\usepackage{listings}
\usepackage[numbers]{natbib}
\usepackage{CJKutf8} 
\usepackage{amsmath}
\usepackage{tcolorbox}
\usepackage{tikz}
\usepackage{pgfplots}
\pgfplotsset{compat=1.18}
\usetikzlibrary{patterns, positioning, calc}
\usepackage{colortbl}
\definecolor{codegreen}{rgb}{0,0.6,0}
\definecolor{codegray}{rgb}{0.5,0.5,0.5}
\definecolor{codepurple}{rgb}{0.58,0,0.82}
\definecolor{backcolour}{rgb}{0.95,0.95,0.92}
\definecolor{promptcolor}{HTML}{D1D0F2}
\definecolor{promptcolorheader}{HTML}{bdbcec}

\definecolor{myblue}{RGB}{135, 170, 255}   
\definecolor{mydarkblue}{RGB}{65, 105, 225} 
\definecolor{mygraydark}{RGB}{169, 169, 169} 
\definecolor{mygraylight}{RGB}{220, 220, 220} 
\newcommand{\promptbox}[2]{
\begin{tcolorbox}[
top=0.3em,bottom=0.3em,left=0.5em,right=0.5em,
toptitle=0.3em,bottomtitle=0.2em,boxsep=0pt,
colframe=promptcolorheader,colback=promptcolor!50,boxrule=0.5pt,
]
\footnotesize
\end{tcolorbox}
}
\lstdefinestyle{mystyle}{
    backgroundcolor=\color{backcolour},   
    commentstyle=\color{codegreen},
    keywordstyle=\color{magenta},
    numberstyle=\tiny\color{codegray},
    stringstyle=\color{codepurple},
    basicstyle=\ttfamily\footnotesize,
    breakatwhitespace=false,         
    breaklines=true,                 
    captionpos=b,                    
    keepspaces=true,                 
    numbers=left,                    
    numbersep=5pt,                  
    showspaces=false,                
    showstringspaces=false,
    showtabs=false,                  
    tabsize=2
}

\title{
    MMFineReason: Closing the Multimodal Reasoning Gap via Open Data-Centric Methods
}

\author[1,2]{Honglin Lin$^*$}
\author[1,3]{Zheng Liu$^*$}
\author[1]{Yun Zhu$^*$}
\author[1,4]{Chonghan Qin}
\author[1]{Juekai Lin}
\author[1]{Xiaoran Shang}
\author[1]{Conghui He}
\author[3]{Wentao Zhang}
\author[1\dag]{Lijun Wu}

\affiliation[1]{Shanghai Artificial Intelligence Laboratory, OpenDataLab}
\affiliation[2]{Shanghai Jiao Tong University}
\affiliation[3]{Peking University}
\affiliation[4]{The University of Hong Kong}

\newtcolorbox{findbox}[2][]{
    enhanced,
    colback=teal!7!white,      
    colframe=teal!80!black,    
    arc=3pt,
    title=#2,
    fonttitle=\bfseries\large,
    coltitle=white,
    colbacktitle=teal!80!black,
    attach boxed title to top center={yshift=-2mm},
    boxed title style={
        colframe=teal!80!black,
        arc=3pt,
    },
    coltext=black,
    fontupper=\linespread{1.2}\selectfont,
    #1
}
\newcommand{\myfind}[3][]{%
    \begin{findbox}[#1]{#2}
    #3
    \end{findbox}%
}

\abstract{
Recent advances in Vision Language Models (VLMs) have driven significant progress in visual reasoning. However, open-source multimodal models still lag behind proprietary systems, largely due to the lack of high-quality reasoning data. Existing datasets offer limited coverage of challenging domains such as STEM diagrams and visual puzzles, and lack consistent, long-form Chain-of-Thought (CoT) annotations essential for eliciting strong reasoning capabilities.
To bridge this gap, we introduce MMFineReason, a large-scale multimodal reasoning dataset comprising 1.8M samples and 5.1B solution tokens, featuring high-quality reasoning annotations distilled from Qwen3-VL-235B-A22B-Thinking. 
The dataset is established via a systematic three-stage pipeline: (1) large-scale data collection and standardization, (2) CoT rationale generation, and (3) comprehensive selection based on reasoning quality and difficulty awareness.
The resulting dataset spans STEM problems, visual puzzles, games, and complex diagrams, with each sample annotated with detailed, visually grounded reasoning traces.
We fine-tune Qwen3-VL-Instruct on MMFineReason to develop MMFineReason-2B/4B/8B versions. Our models establish new state-of-the-art (SOTA) results for their size class. Notably, MMFineReason-4B succesfully surpasses Qwen3-VL-8B-Thinking, and MMFineReason-8B even outperforms Qwen3-VL-30B-A3B-Thinking while approaching Qwen3-VL-32B-Thinking, demonstrating remarkable parameter efficiency.
Crucially, we uncover a "less is more" phenomenon via our difficulty-aware filtering strategy: a subset of just 7\% (123K samples) achieves performance comparable to the full dataset.
Notably, we reveal a synergistic effect where reasoning-oriented data composition simultaneously boosts general capabilities.
Further, we conduct comprehensive ablation studies on training strategies and data composition, providing key insights and practical recipes for multimodal reasoning model development. 
For open-source, we release the full dataset and models to facilitate reproducible research on data-centric strategies for multimodal reasoning.
}

\date{\today}
\metadata[Equal Contributions]{Honglin Lin, Zheng Liu, Yun Zhu}
\correspondence{Lijun Wu, \email{lijunwu@pjlab.org.cn}}
\metadata[Homepage]{\url{https://mmfinereason.github.io/}}
\metadata[HuggingFace]{\url{https://huggingface.co/collections/OpenDataArena/mmfinereason}}

\begin{document}

\maketitle

\definecolor{myblue}{RGB}{135, 170, 255}        
\definecolor{mymediumgray}{RGB}{169, 169, 169}  
\definecolor{mydeepgray}{RGB}{105, 105, 105}    
\definecolor{mylightgray}{RGB}{225, 225, 225}   

\begin{figure}[t]
    \centering
    \includegraphics[width=1\linewidth]{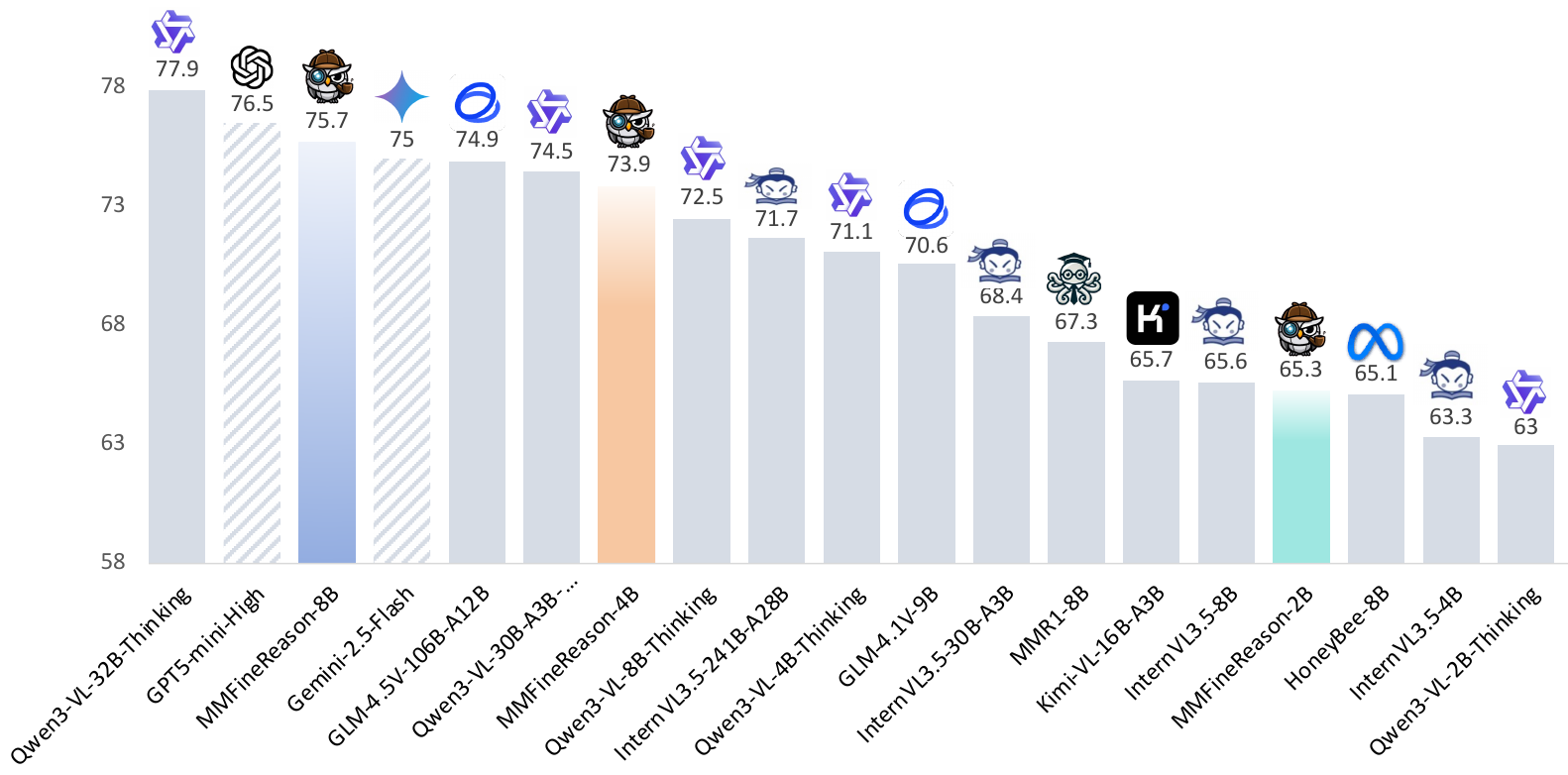}
    \caption{Average score across mathematical reasoning and multimodal understanding benchmarks. MMFineReason-2B/4B/8B demonstrates strong performance relative to thinking models with significantly more parameters.}
    \label{fig:compare_result}
\end{figure}

\section{Introduction}
\label{section:intro}
Recent advances in Vision Language Models (VLMs) have led to substantial improvements in visual reasoning capabilities~\cite{lmm_survey, reason_survey}. State-of-the-art (SOTA) proprietary systems such as GPT-5~\cite{gpt5} and Gemini 3~\cite{gemini3} achieve strong performance on complex multimodal tasks, benefiting from access to massive, carefully curated private datasets. Mounting evidence suggests that  the quality and structure of training data plays a central role in unlocking strong reasoning abilities~\cite{data_survey}.

In the text domain, this data-centric paradigm has proven highly effective. The release of DeepSeek-R1~\cite{deepseekr1} and its distilled variants have further highlighted that constructing high-quality post-training data is key to closing the gap with proprietary models. Recent works~\cite{openr1, openthoughts, limo, caco, liu2026chartversescalingchartreasoning,  niu2025mineru25decoupledvisionlanguagemodel,metaladder} have extensively investigated pipelines for building high-quality training data, enabling open models to approach proprietary performance.

However, extending this success to the multimodal setting remains challenging. The open-source community still struggles to produce multimodal datasets that match the scale, consistency, and reasoning depth of those used by leading proprietary models. Although recent efforts such as FineVision~\cite{finevision} and LLaVA-OneVision-1.5~\cite{llavaov-1.5} have expanded both the quality and quantity of available data, two critical limitations persist:

\begin{itemize}
    \item \textbf{Data imbalance}: While VQA data derived from natural images and documents is largely sufficient, high-quality visual reasoning samples—particularly for STEM diagrams and visual puzzles—remain scarce due to inherent data rarity and high annotation costs~\cite{lin2026scientific}.
    
    \item \textbf{Inconsistent reasoning quality}: 
    Unlike the text domain, where distillation from strong teacher models such as DeepSeek-R1 has become a standard practice for obtaining high-quality rationales. Multimodal datasets remain fragmented and heterogeneous in annotation style, with limited availability of interpretable, long-form Chain-of-Thought (CoT) supervision~\cite{wemath,mmk12}.
\end{itemize}

These limitations hinder principled, data-centric research on multimodal reasoning. A natural solution is to distill reasoning traces from a capable multimodal teacher. The recently released Qwen3-VL series~\cite{qwen3vl} demonstrates strong visual reasoning capabilities approaching proprietary systems, making it a promising candidate for scalable data synthesis.

\begin{figure}[t]
    \centering
    \includegraphics[width=1\linewidth]{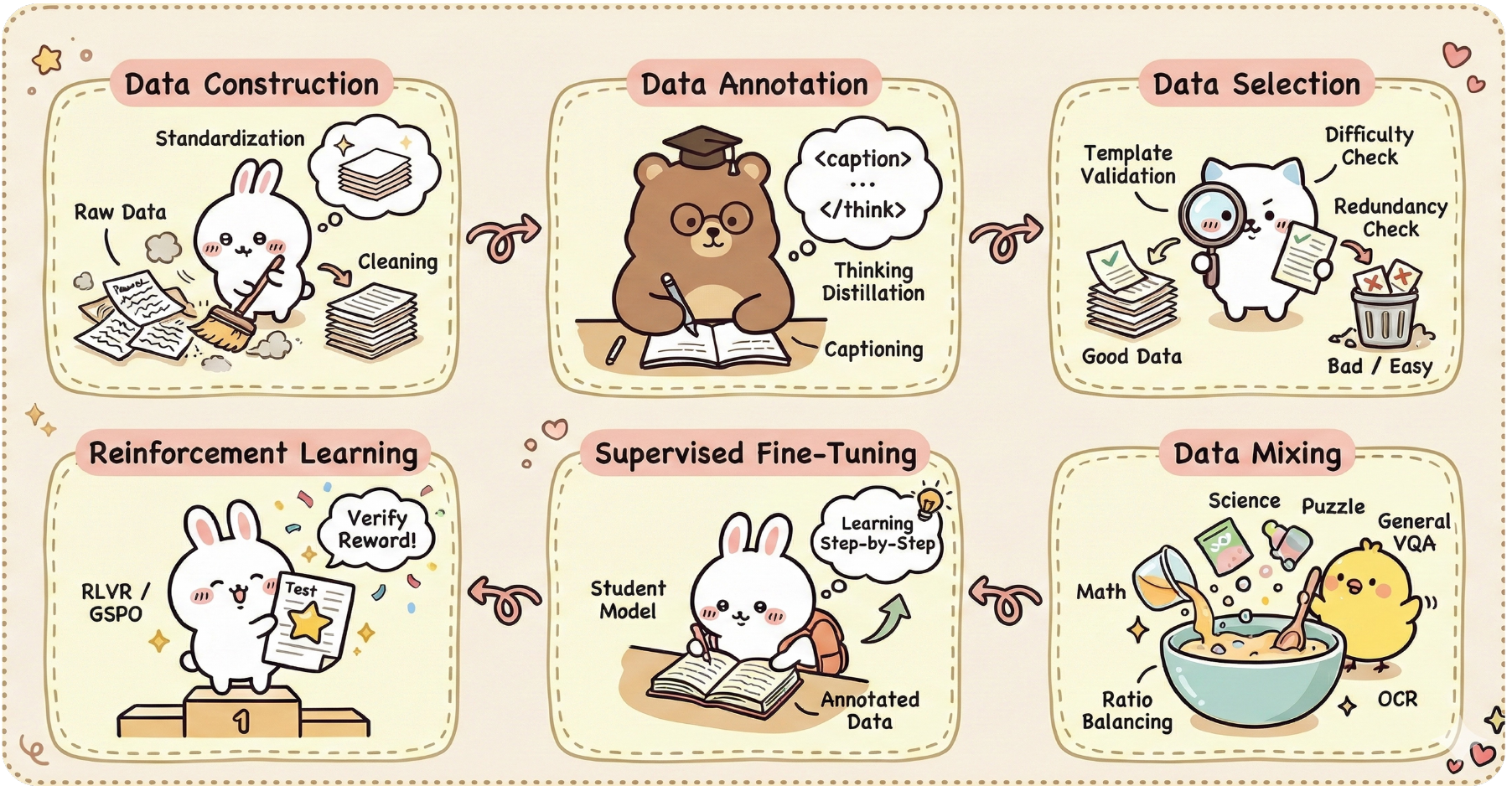}
    \caption{MMFineReason data pipeline and the two-stage training. Illustrating data construction, annotation, selection, mixing, and model training (SFT and RL) in our framework.}
    \label{fig:pipeline}
\end{figure}

Motivated by this, we introduce \textbf{MMFineReason}, an open-source dataset of over 1.8M samples with 5.1B solution tokens, featuring high-quality reasoning annotations distilled from Qwen3-VL-235B-A22B-Thinking. As illustrated in Figure~\ref{fig:pipeline}, 
we build the dataset in three stages: 
(1) Data Aggregation and Standardization, where we collect, clean, and unify raw multimodal data from diverse sources into a canonical schema; 
(2) Reasoning Distillation, where we generate detailed, visually grounded reasoning traces using the SOTA teacher model; and 
(3) Data Selection, where we conduct rigorous quality verification and difficulty-based filtering to ensure both correctness and training efficiency.

By fine-tuning Qwen3-VL-Instruct on our constructed MMFineReason dataset, we obtain MMFineReason-2B/4B/8B models with superior performance. As shown in Figure~\ref{fig:compare_result}, our models establish a new SOTA among open-source models of comparable size. Notably, MMFineReason-4B can surpass Qwen3-VL-8B-Thinking, and MMFineReason-8B even outperforms Qwen3-VL-30B-A3B-Thinking while approaching Qwen3-VL-32B-Thinking.

Our contributions can be summarized as follows:

\begin{itemize}
    \item We design a systematic pipeline integrating data collection, filtering, and CoT distillation to construct MMFineReason-1.8M, the first large-scale multimodal reasoning dataset comprising 5.1B tokens with high-quality annotations distilled from Qwen3-VL-235B-A22B-Thinking.
    \item We develop the MMFineReason model family (2B/4B/8B) by fine-tuning Qwen3-VL-Instruct on this dataset. Notably, our models exhibit exceptional scaling efficiency: MMFineReason-4B surpasses Qwen3-VL-8B-Thinking, while MMFineReason-8B outperforms Qwen3-VL-30B-A3B-Thinking and approaches the performance of Qwen3-VL-32B-Thinking.
    \item We demonstrate that training with reasoning-oriented data, such as STEM, puzzles, games, diagrams, yields a synergistic effect, driving simultaneous improvements in both specialized reasoning capabilities and general model performance.
    \item We adopt a difficulty-aware filtering strategy to construct efficient fine-tuning subsets, specifically MMFineReason-123K and MMFineReason-586K. Remarkably, fine-tuning on just 7\% of the data selected achieves performance comparable to training on the full dataset.
    \item We conduct comprehensive ablation studies on training strategies and data composition, providing key insights and practical recipes for multimodal reasoning model development.
\end{itemize}
In summary, MMFineReason serves as a high-quality training resource and a reproducible framework for building open-source VLMs with robust reasoning capabilities, fostering principled exploration of data-centric strategies in multimodal reasoning.

\myfind{Overall Findings}{
\begin{itemize}
    \item \textbf{Data-centric strategies exhibit strong scaling efficiency.} 
    With a strong pretrained backbone, training on only 5.1B tokens is sufficient to surpass larger open-source models (e.g., Qwen3-VL-30B-A3B-thinking) and achieve performance comparable to advanced closed-source models (e.g., GPT-5-mini-high).

    \item \textbf{A severe cross-domain imbalance in data distribution and difficulty.} 
    A large portion of the current training data is overly simple (67\%), while puzzle-style problems are significantly harder and remain underrepresented compared to STEM reasoning data.

    \item \textbf{Difficulty-aware filtering is highly effective.} 
    The difficulty-filtered MMFineReason-123K dataset achieves performance close to the full dataset using only 7\% of the data, demonstrating substantial data efficiency gains.

    \item \textbf{A reasoning-oriented data mixing strategy enables simultaneous gains in general and reasoning tasks.}
    Only a few amount of general data improves both general-purpose tasks and reasoning benchmarks in a synchronized manner.

    \item \textbf{Ultra-high resolution offers diminishing returns for reasoning.} 
    Extremely large input resolutions (e.g., $2048^2$) bring limited benefits for reasoning tasks, while general vision tasks still require higher resolutions.

    \item \textbf{Caption augmentation provides marginal benefit once chains-of-thought are mature.} 
    When the reasoning chain is already well-formed, concatenating an additional caption brings almost no further improvement.
\end{itemize}
}

\section{Related Work}
\subsection{Multimodal Reasoning Datasets}

Multimodal reasoning---including mathematical problem solving, visual logical reasoning, and chart understanding---is a key challenge for evaluating the reasoning capabilities of vision-language models. Prior work has demonstrated that data is a central factor driving the advancement of model reasoning ability~\cite{reason_survey,openthoughts}. However, in the multimodal domain, data acquisition and synthesis are considerably more difficult~\cite{finevision}. Many proprietary models rely on large-scale private datasets~\cite{gemini3,gpt5,qwen3vl}, resulting in a significant data gap. In contrast, the open-source community still lacks multimodal reasoning datasets of substantial scale.

To support VL reasoning, several datasets have been proposed, such as MathV360K~\cite{mathllava} and LLaVA-CoT~\cite{llava-cot}. However, these datasets primarily focus on mathematical reasoning and therefore provide limited coverage. Although FineVision~\cite{finevision} attempts a large-scale data aggregation, its content is relatively coarse, includes low-quality data sources, and omits many reasoning-oriented datasets. To address these limitations, we conduct the most extensive collection and integration of existing multimodal reasoning data to date, and further provide high-quality reference answers to support research in the open-source community.

\subsection{Data Recipes for Reasoning Models}

Early data pipelines such as LLaVA-Instruct-150K~\cite{llava} used GPT-4~\cite{gpt4} to generate VQA pairs. ShareGPT4V~\cite{sharegpt4v} leveraged GPT-4V~\cite{gpt4v} and ShareCaptioner to produce large-scale image descriptions, with length- and content-based filtering to ensure data quality. Other approaches, such as SynthVLM~\cite{liu2025synthvlmhighqualityefficientsynthesis} and FUSION~\cite{liu2025fusionfullyintegrationvisionlanguage}, further scale up existing datasets by generating synthetic images.
Parallel to advancements in the visual domain, significant progress has been made in constructing text-only reasoning data~\cite{lemma,mathfusion,metaladder}.
With the emergence of DeepSeek-R1~\cite{deepseekr1} and its high-performance distilled models, recent efforts such as OpenR1~\cite{openr1} and OpenThoughts~\cite{openthoughts} have built open and transparent data construction pipelines, enabling the open-source community to approach the capability levels of proprietary systems using only public data~\cite{amthinking}.

Despite this progress, the multimodal field still lacks transparent and reproducible data curation and training pipelines capable of achieving performance parity with closed-source systems. To this end, we introduce MMFineReason, a high-quality and fully reproducible data construction pipeline that helps open-source multimodal models progressively narrow the performance gap with closed-source systems at a reasonable cost. \textbf{Importantly, our data construction workflow is entirely based on locally deployed open-source models and does not rely on any closed-source APIs.}

\section{MMFineReason Pipeline}
In this section, we detail our dataset construction pipeline. First, we describe the data collection and processing procedures in Section~\ref{subsec:datacollect}. Next, we explain our method for distilling responses for curated datasets in Section~\ref{subsec:distill}. Finally, we present the data selection strategies in Section~\ref{subsec:dataselect}.

\subsection{Data Collection and Processing}\label{subsec:datacollect}
\paragraph{Data Collection.}

We initiate our data construction by aggregating a diverse array of multimodal datasets from the open-source community. We first leverage FineVision~\cite{finevision}, an extensive collection of visual instruction datasets. We conduct a rigorous manual inspection of each candidate source, filtering out datasets unrelated to STEM or reasoning tasks. To expand the coverage of mathematical reasoning, scientific reasoning, and visual games \& puzzles, we further incorporate high-quality datasets such as BMMR~\cite{bmmr}, Euclid30K~\cite{euclid}, Zebra-CoT-Physics~\cite{zebracot}, and GameQA-140K~\cite{gameqa}. This strategic expansion ensures a more balanced and challenging coverage of scientific and puzzle-solving domains. For comprehensive details and inclusion criteria, please refer to Appendix~\ref{sec:subset}.

\paragraph{Data Cleaning.}

We implement a comprehensive data cleaning pipeline to guarantee the linguistic consistency, textual cleanliness, and reasoning suitability of the collected samples. The cleaning prompt is detailed in Prompt~\ref{pmt:clean} (see Appendix~\ref{apx:case} for representative noise cases). The procedure includes:

\begin{itemize}
    \item \textbf{Language Standardization}: To unify the linguistic landscape of the corpus, we translate non-English questions found in datasets like \texttt{BMMR} and \texttt{Euclid30K} into English.
    \item \textbf{Noise Removal}: We sanitize question text by removing extraneous artifacts, including webpage links, corrupted characters, formatting residues, problem indices, and score annotations.
    \item \textbf{Instruction Refinement}: We reformulate prompts that solicit shallow responses (e.g., ``directly give the answer'') into directives that explicitly encourage analytical thinking (e.g., ``provide your answer after careful reasoning''). This step is crucial for eliciting high-quality reasoning traces during distillation.
    \item \textbf{Task Suitability Filtering}: We filter out tasks that fall outside the scope of visual analytical reasoning, such as coding exercises or generation-based drawing tasks, ensuring the dataset remains focused on logical problem-solving.
\end{itemize}

We further perform automatic image cleaning~\cite{finevision} by discarding corrupted or unreadable images, resizing those with a longest side exceeding 2048 pixels while preserving aspect ratio, and converting all images uniformly to the RGB color space.

\paragraph{Data Standardization.}
\label{sec:standardization}
The landscape of multimodal datasets is characterized by significant fragmentation and a lack of standardization. The heterogeneity in file formats and annotation structures across different sources poses a substantial challenge for unified data processing. For datasets that lack explicit ground-truth labels, we employ \texttt{Qwen3-30B-A3B-Thinking} to extract and store the missing answers in the \texttt{answer} field, following the prompt in Prompt~\ref{pmt:answer_extraction}. To facilitate downstream processes—such as caption generation, reasoning distillation, and correctness evaluation—and to enhance accessibility for the research community, we further convert all collected samples into a unified canonical schema. Each standardized data entry includes the following fields:
\begin{itemize}
    \item \textbf{Metadata}: \texttt{source}, \texttt{id}
    \item \textbf{Raw Data}: \texttt{original\_question}, \texttt{original\_answer}
    \item \textbf{Input/Output}: \texttt{image}, \texttt{question}, \texttt{answer}
    \item \textbf{Augmented Annotations}: \texttt{qwen3vl\_235b\_instruct\_caption}, \texttt{qwen3vl\_235b\_thinking\_response}
    \item \textbf{Metrics}: \texttt{qwen3vl\_4b\_pass\_rate}, \texttt{is\_consistent}, \texttt{consistency\_analysis}
\end{itemize}
\noindent
The schema fields are defined as follows: \texttt{source} denotes the origin dataset (e.g., Geometry3K), while \texttt{id} serves as the unique sample identifier. The Raw Data fields preserve the \texttt{original\_question} and \texttt{original\_answer} exactly as obtained from the source, which are then processed into the standardized Input/Output triplet (\texttt{image}, \texttt{question}, and \texttt{answer}) used for training.

Within the Augmented Annotations, \texttt{qwen3vl\_235b\_instruct\_caption} and \texttt{qwen3vl\_235b\_thinking\_response} denote the dense visual description and reasoning steps generated by the teacher model, respectively. The Metrics group includes \texttt{qwen3vl\_4b\_pass\_rate}, which serves as a difficulty proxy based on a smaller model's performance, alongside \texttt{is\_consistent} and \texttt{consistency\_analysis}, which provide automated verification of the generated reasoning against the ground truth.

\subsection{Data Annotation}\label{subsec:distill}

We employ \texttt{Qwen3-VL-235B-A22B-Thinking}, currently recognized as the SOTA open-source VLM, to distill long-form CoT explanations for each sample.
To ensure the rigor and reproducibility of the reasoning process, the distillation prompt (see Prompt~\ref{pmt:distill}) imposes a systematic four-phase solution framework: \textit{Comprehensive Information Extraction}, \textit{Strategic Problem Setup}, \textit{Rigorous Solution Execution}, and \textit{Solution Validation}. Furthermore, it explicitly instructs the model to treat visual elements as integral components of the solution rather than supplementary context.

The prompt also enforces a unified output template: the model first emits a multi-step reasoning trace wrapped in a \verb|<think>...</think>| block, followed by the final solution wrapped in an \verb|<answer>...</answer>| block to facilitate downstream answer parsing and automatic verification. Additionally, we utilize \texttt{Qwen3-VL-235B-A22B-Instruct} to generate dense image captions, following the guidelines in Prompt~\ref{pmt:caption}.

Through this systematic annotation and distillation process, we obtain the original MMFineReason-Full dataset, consisting of \textbf{2.3M} samples with a total of \textbf{8.8B} solution tokens. 
A detailed breakdown of the dataset composition is provided in Table~\ref{tab:data_stats_split}.

\subsection{Data Selection}\label{subsec:dataselect}

\begin{figure}[h]
    \centering
\includegraphics[width=1\linewidth]{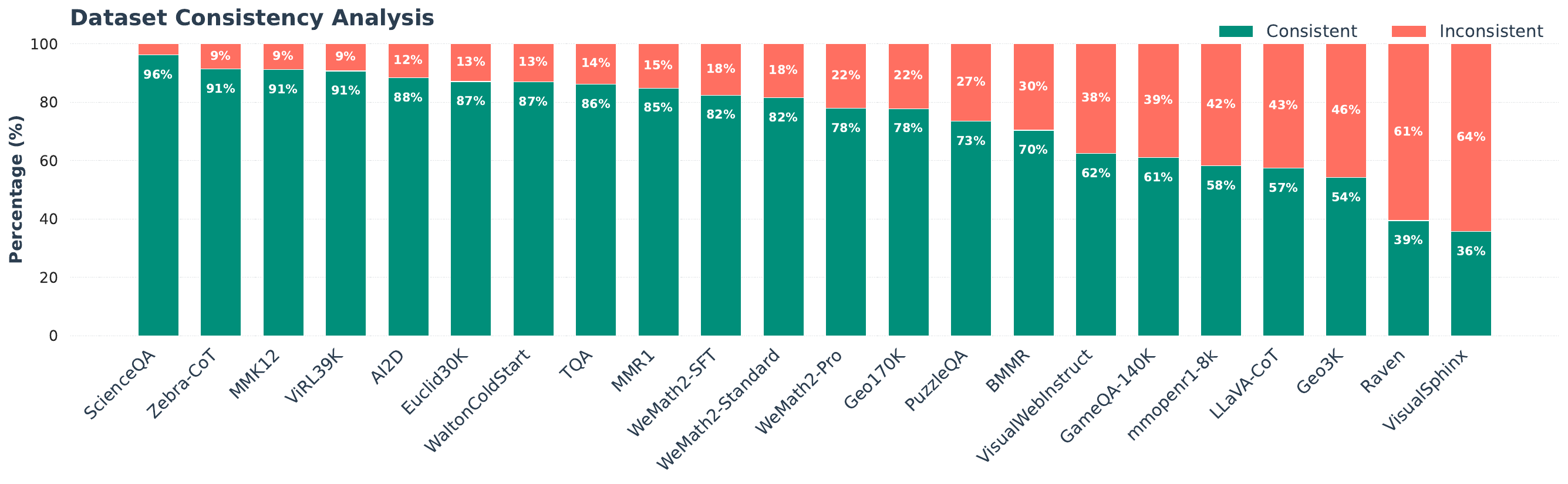}
    \caption{Consistency analysis across visual instruction tuning datasets. The chart displays the ratio of samples where the predictions generated by Qwen3-VL-235-A22B-Thinking align with the original ground truth answers ("Consistent") versus cases of disagreement ("Inconsistent").}
    \label{fig:consistent}
\end{figure}

\paragraph{Reasoning Quality Filtering to Construct MMFineReason-1.8M.}
We adopt a simple and light way to construct our MMFineReason dataset for training. Specifically, to ensure high-quality and non-redundant reasoning traces, we apply a multi-stage filtering procedure to the distilled outputs.

\begin{itemize}
    \item \textbf{Template and Length Validation}: We first impose strict structural validation to ensure the usability of the distilled output. Specifically, we filter out any reasoning trace that fails to adhere to the mandated \verb|<think>...</think>| and \verb|<answer>...</answer>| output template. Furthermore, to prevent the retention of superficial or trivial rationales, we enforce a minimum length constraint, discarding traces that are shorter than 100 words. This stage removes approximately 1.2\% of the data based on length and template constraints (Table~\ref{app:datacleaning}). The detailed statistics of the processed subsets are reported in Table~\ref{app:datacleaning}.
    \item \textbf{N-gram De-duplication}: We detect and remove templated or overly repetitive CoTs using an n-gram overlap criterion. Concretely, we flag CoTs that contain any 50-gram that repeats at least 3 times (i.e., $n=50$, frequency threshold $f=3$). Flagged traces are either discarded or re-generated with a different random seed to encourage diversity.
    \item \textbf{Correctness Verification}: For tasks that have ground-truth answers, we extract the final answer from the \verb|<answer>| tag and compare it against the answer extracted in Section~\ref{sec:standardization}, the incorrect CoTs are discarded. The detailed verification protocol is provided in Prompt~\ref{pmt:verify}. This process eliminated roughly 20\% of instances containing potential hallucinations or incorrect reasoning traces, the consistency ratios across different subsets are summarized in Figure~\ref{fig:consistent}. The detailed verification results are shown in Table~\ref{tab:pass_rate_stats}.
\end{itemize}

Following the above data selection pipeline, we obtain a high-quality \textbf{MMFineReason-1.8M} dataset with totally \textbf{5.1B} solution tokens. 
We further uniformly sample 40k instances from MMFineReason-1.8M to construct an RL training subset, while the remaining data are reserved for SFT.

\paragraph{Difficulty Filtering for Efficient Training.}\label{para:difficulty}
Given the massive scale of our curated data, training on the entire corpus is computationally suboptimal due to the prevalence of redundant or trivial samples. To address this, we employ a difficulty-based filtering strategy~\cite{limo}. Specifically, we perform inference on each question using \texttt{Qwen3-VL-4B-Thinking}, generating four independent responses. We discard any example where the model provides a correct answer in at least one attempt (pass rate = 0). This conservative filtering criterion ensures that we retain only genuinely challenging problems where a moderate-sized model fails consistently. By discarding samples that contribute negligible training signals, we direct our computational resources toward challenging data points that actively drive optimization, resulting in faster convergence. 
Specifically, we derive two challenging subsets from the full MMFineReason-1.8M corpus: \textbf{MMFineReason-123K} and \textbf{MMFineReason-586K}, by retaining samples with \textbf{pass rate = 0} and \textbf{pass rate $\neq$ 1}, respectively, which are well suited for efficient SFT and ablation studies.
A detailed analysis of the difficulty score distribution is presented in Section~\ref{subsec:filter}.

\section{MMFineReason Dataset}
In this section, we present a comprehensive analysis of MMFineReason. We begin by delineating the dataset composition in Section~\ref{subsec:composition}. Subsequently, we investigate visual diversity in Section~\ref{subsec:image}, with a specific focus on image category distribution and visual granularity. Finally, Section~\ref{subsec:response} provides a statistical analysis of response characteristics and benchmarks our dataset against existing multi-modal reasoning datasets.
\begin{figure}[t]
    \centering
    \includegraphics[width=1\linewidth]{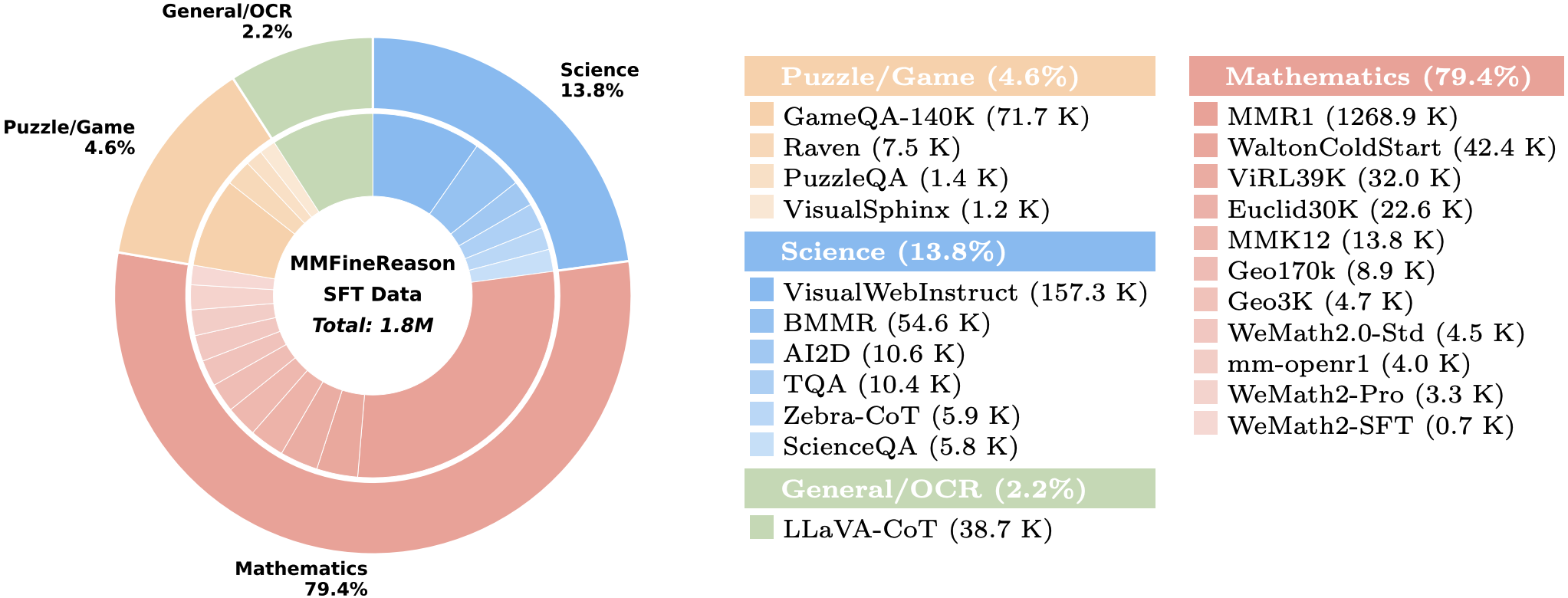}
    \caption{Dataset composition of MMFineReason-1.8M. The outer ring represents the proportion of major categories, and the inner ring shows the distribution of specific datasets. \textbf{Note:} To ensure the visual legibility of diverse domains, the segment sizes in this chart are scaled by the \textbf{square root} of sample counts ($\sqrt{N}$). The actual data distribution is dominated by Mathematics (79.4\%), followed by Science (13.8\%), Puzzle/Game (4.6\%), and General/OCR (2.2\%).}
    \label{fig:dataset-composition}
\end{figure}
\subsection{Dataset Composition}\label{subsec:composition}

MMFineReason contains approximately 1.8 million (specifically 1,770,926) high-quality multimodal reasoning samples. Figure~\ref{fig:dataset-composition} summarizes the domain distribution of the dataset, which is strategically weighted towards symbolic and logic-heavy tasks: Mathematics (79.4\%), Science (13.8\%), Puzzle/Game (4.6\%), and General/OCR (2.2\%).

\textbf{Mathematics (79.4\%).} This domain forms the backbone of our reasoning supervision, primarily sourced from the massive MMR1~\cite{mmr1} dataset (1.27M). To ensure diversity in problem types, we integrate WaltonColdStart~\cite{walton} (42.4K) and ViRL39K~\cite{virl} (32.0K). We further enhance geometric and symbolic reasoning capabilities by including Euclid30K~\cite{euclid} (22.6K), MMK12~\cite{mmk12} (13.8K), Geo170K~\cite{finevision} (8.9K), and Geo3K~\cite{finevision} (4.7K). Finally, we incorporate specialized subsets including mm-openr1~\cite{mmopenr1} (4.0K) and the comprehensive WeMath family~\cite{wemath} (Standard 4.5K, Pro 3.3K, and SFT 0.7K).

\textbf{Science (13.8\%).} Science constitutes a significant portion of the dataset, anchored by VisualWebInstruct~\cite{finevision} (157.3K) and BMMR~\cite{bmmr} (54.6K). These are complemented by smaller, high-quality collections such as TQA~\cite{finevision} (10.4K) and AI2D~\cite{finevision} (10.6K), along with domain-specific subsets like Zebra-CoT~\cite{zebracot} (5.9K) and ScienceQA~\cite{finevision} (5.8K).

\textbf{Puzzle/Game (4.6\%).} This domain targets strategic planning and abstract pattern recognition. It is dominated by GameQA-140K~\cite{gameqa} (71.7K) and further enriched by Raven~\cite{finevision} (7.5K), VisualSphinx~\cite{visualsphinx} (1.2K), and PuzzleQA~\cite{puzzlevqa} (1.4K).

\textbf{General/OCR (2.2\%).} In contrast to general-data-heavy training recipes, we adopt a reasoning-dominant composition. Empirically, the base model's visual perception is already robust, and extensive general data often yields diminishing returns for reasoning tasks. Therefore, we include only 38.7K general-purpose samples from LLaVA-CoT~\cite{llava-cot}, serving as a regularization set to preserve broad visual and OCR capabilities without diluting the reasoning-centric supervision.

\begin{figure}[t!]
    \centering
    \includegraphics[width=1\linewidth]{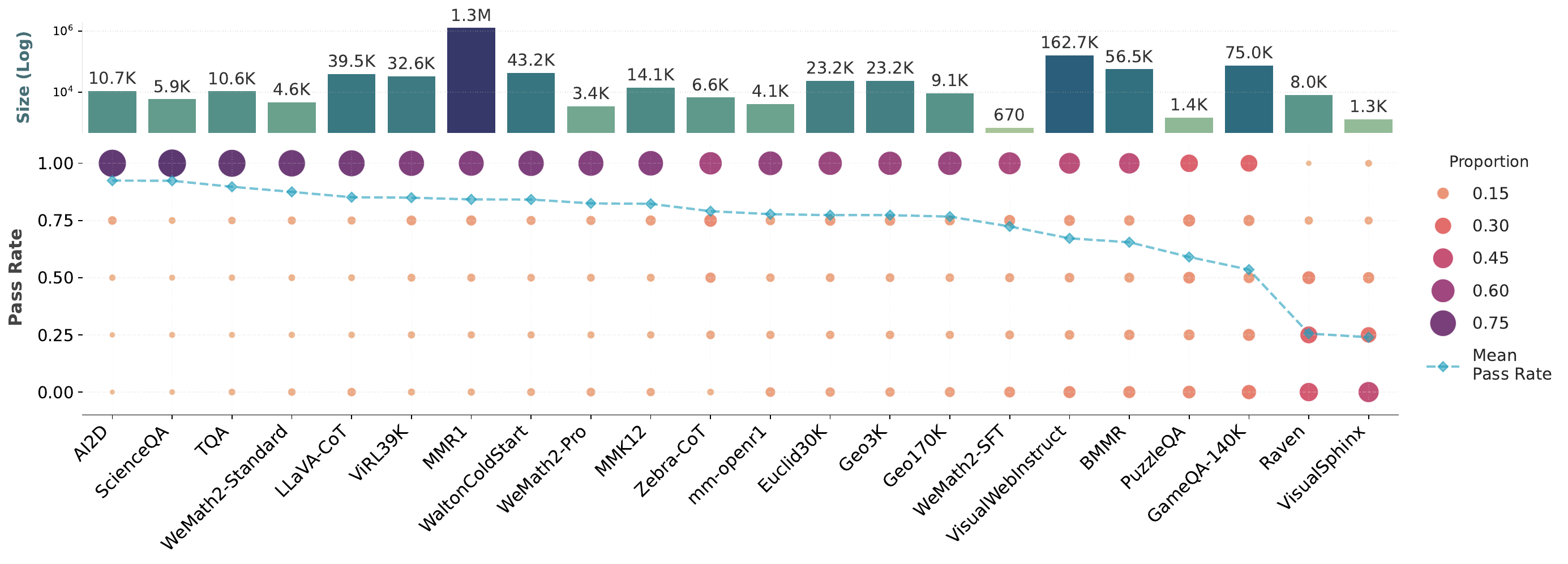}
    \caption{Pass rate distribution across sub-datasets. Datasets are sorted by descending mean pass rate (easiest to hardest). The bubble chart encodes sample proportion via size and color intensity, overlaid with a mean pass rate trendline.}
    \label{fig:difficulty_dist}
\end{figure}

\subsection{Difficulty Distribution Analysis}\label{subsec:filter}

Building on the filtration technique for efficient training described in Section~\ref{para:difficulty}, Figure~\ref{fig:difficulty_dist} illustrates the pass rate distribution across various sub-datasets using a bubble chart overlaid with a mean pass rate line. The datasets are arranged on the x-axis by mean pass rate in descending order (left to right), ensuring a visual progression from easiest to hardest, while the y-axis represents the ``Pass Rate'' from 0.00 (hardest) to 1.00 (easiest). Bubble size and color intensity denote the proportion of samples at a specific pass rate level.

Notably, science-oriented sub-datasets such as ScienceQA, AI2D, and TQA exhibit relatively high pass rates. These datasets are generally considered simpler because they feature clean, synthetic diagrams and primarily rely on knowledge derived from primary and secondary school textbooks. Furthermore, they lack the visual complexity and expert depth required by modern standards and are predominantly Multiple Choice Questions, which limits the solution space. Conversely, puzzle and game-based datasets like GameQA-140K, Raven, and VisualSphinx demonstrate the lowest pass rates. These sub-datasets require multi-step abstract reasoning and fine-grained visual discrimination, resulting in a significant concentration of samples with low pass rates. Furthermore, we observe that the proportion of samples in the intermediate range remains sparse across all sub-datasets. This is because the reasoning process involved often follows a binary success outcome. Unlike tasks where partial understanding might yield closer approximations, these logic and puzzle problems require a rigorous, unbroken chain of deduction; a single failure in any reasoning step or visual grounding typically leads to a completely incorrect answer, thereby polarizing the distribution toward the extremes (0 or 1) and leaving the middle interval empty.

\begin{table}[t!]
    \centering
    \caption{Comparison of token length statistics across datasets. We report the distribution metrics (mean, median, and percentiles) for reasoning chains (CoT) and image captions. The \textbf{bold} values indicate the highest statistics, highlighting the significant reasoning depth of MMFineReason.}
    \label{tab:token-comparison}
    \resizebox{\textwidth}{!}{
    \begin{tabular}{l l r r r r r r r r r r}
        \toprule
        \textbf{dataset} &
        \textbf{type} &
        \textbf{count} &
        \textbf{total\_tokens} &
        \textbf{mean} &
        \textbf{median} &
        \textbf{std\_dev} &
        \textbf{min} &
        \textbf{max} &
        \textbf{p25} &
        \textbf{p75} &
        \textbf{p95} \\
        \midrule
        \textbf{MMFineReason (Ours)} & CoT     & 1770926   & \textbf{5152806394} & \textbf{2909.67} & \textbf{2038} & \textbf{2463.83} & 239 & \textbf{16316} & \textbf{1321} & \textbf{3569} & \textbf{8207} \\
        OpenMMReasoner     & CoT     & 874357    & 590096263  & 674.89  & 180  & 1477.53 & 26  & 16483 & 102 & 464  & 3318  \\
        HoneyBee           & CoT     & \textbf{2481229}   & 2636405079 & 1062.54 & 972  & 428.00  & 203 & 7190  & 745  & 1298 & 1931  \\ \midrule
        \textbf{MMFineReason (Ours)} & Caption & \textbf{1770926}   & \textbf{1079313259} & \textbf{609.46}  & \textbf{582}  & \textbf{184.88}  & 1   & \textbf{5187}  & \textbf{494}  & \textbf{688}  & \textbf{920}   \\
        HoneyBee           & Caption & 1439921   & 431096653  & 299.39  & 264  & 157.99  & 21  & 2739  & 201 & 350  & 598   \\
        \bottomrule
    \end{tabular}
    }
\end{table} 

\begin{figure}[t]
    \centering

    \begin{subfigure}[t]{0.32\linewidth}
        \centering
        \includegraphics[width=\linewidth]{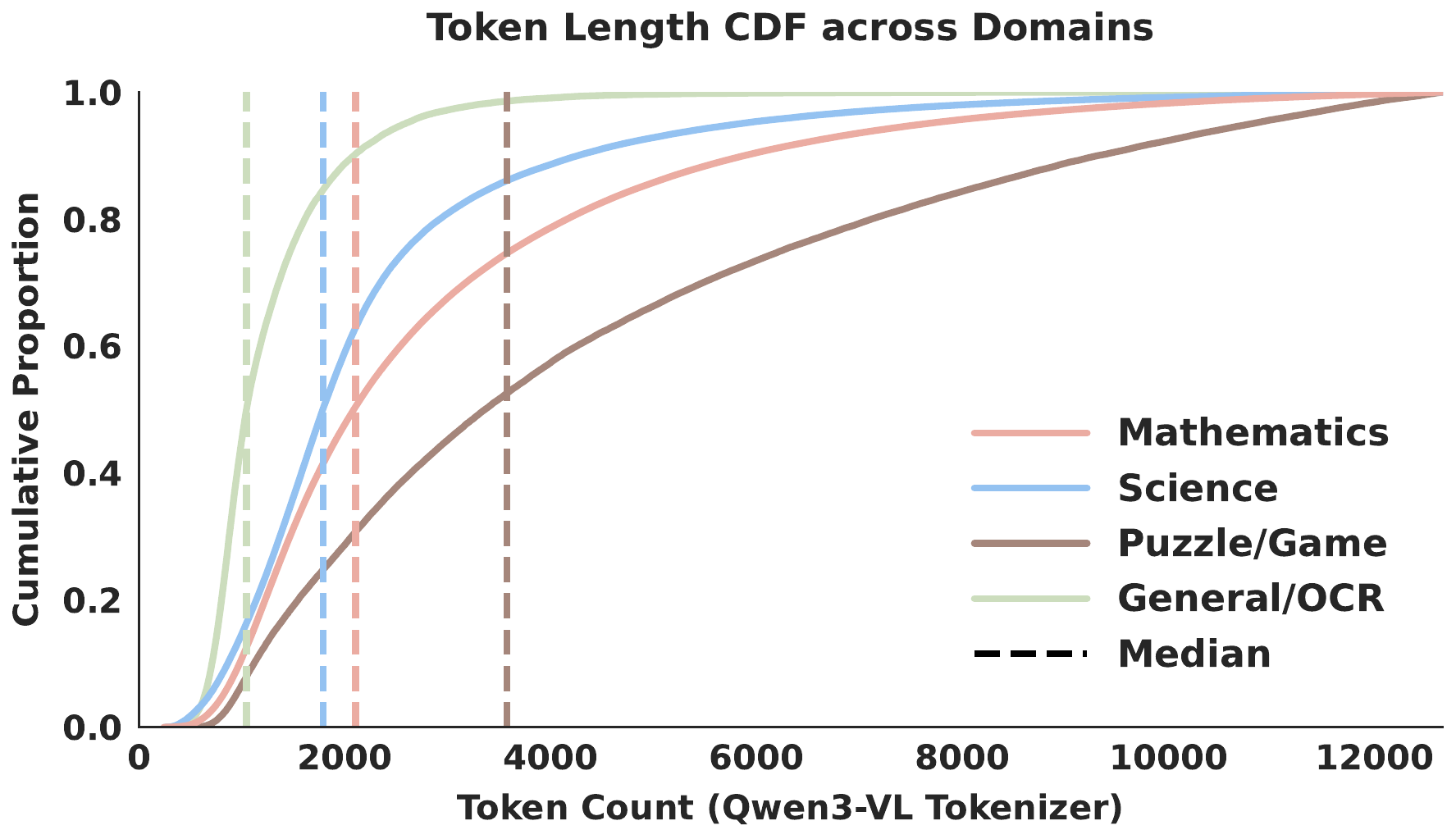}
        \phantomsubcaption
        \label{fig:cot-token-distribution}
    \end{subfigure}
    \hfill
    \begin{subfigure}[t]{0.32\linewidth}
        \centering
        \includegraphics[width=\linewidth]{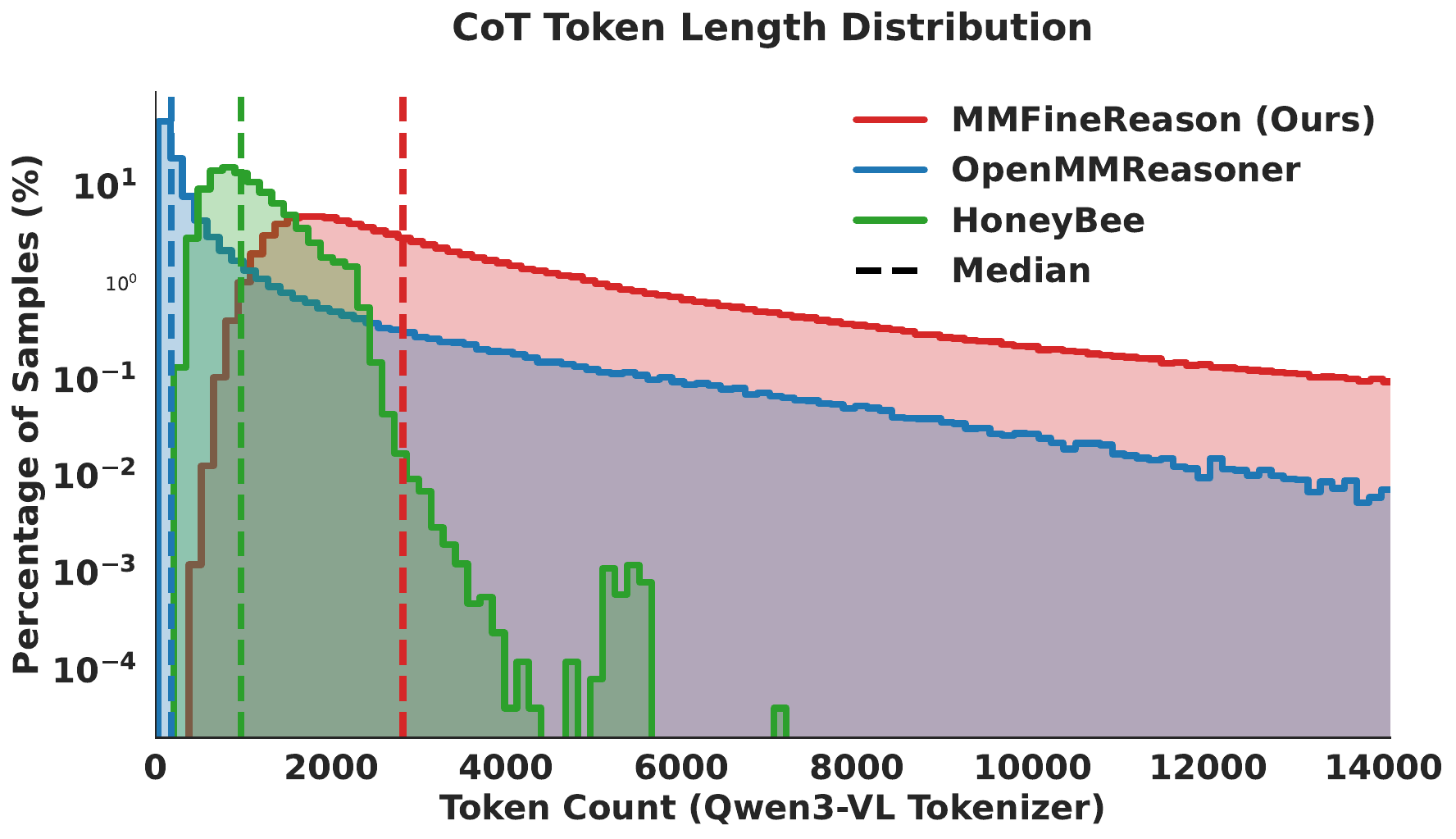}
        \phantomsubcaption
        \label{fig:cot-token-comparison}
    \end{subfigure}
    \hfill
    \begin{subfigure}[t]{0.32\linewidth}
        \centering
        \includegraphics[width=\linewidth]{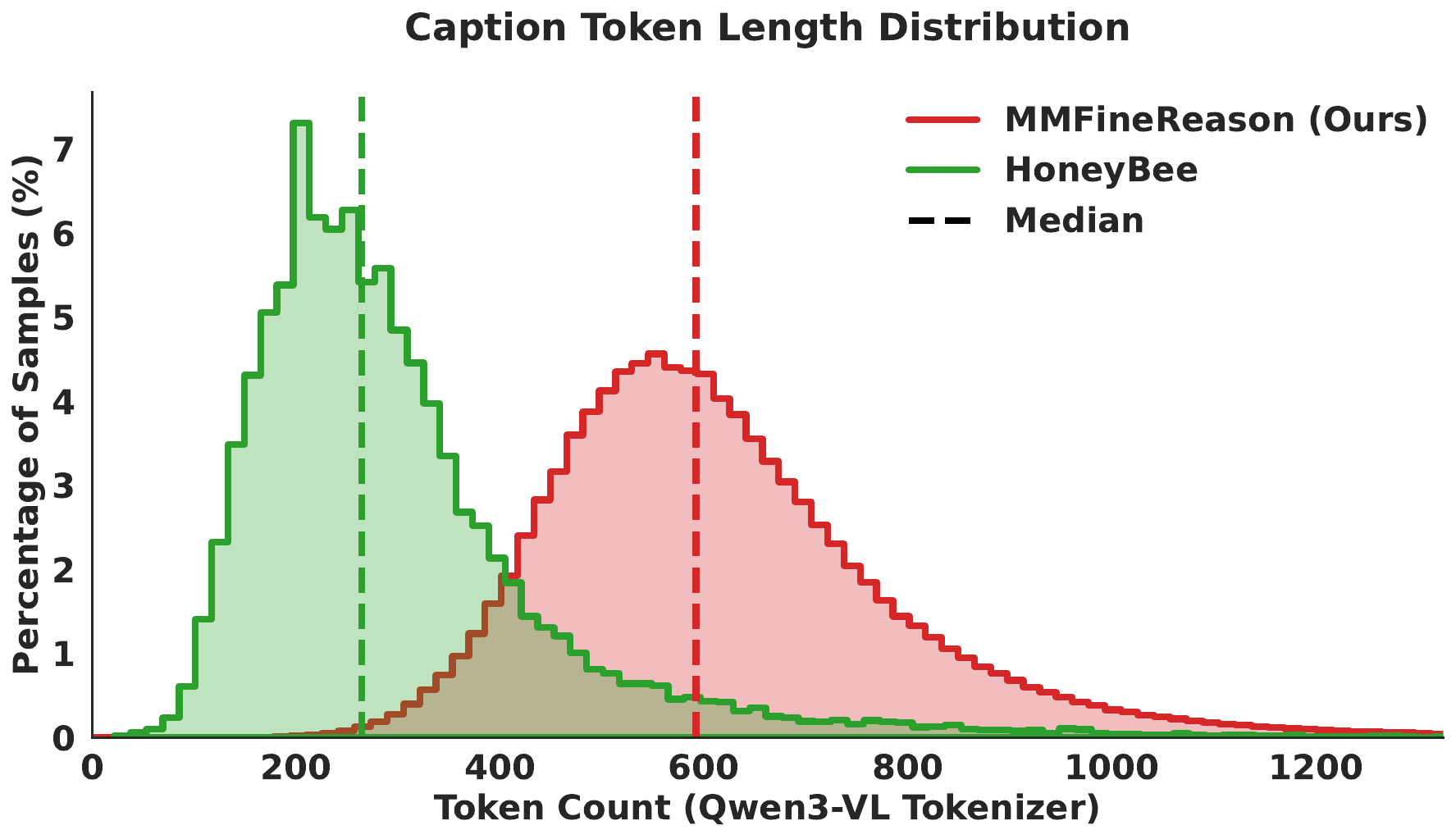}
        \phantomsubcaption
        \label{fig:caption-token-comparison}
    \end{subfigure}

    \caption{Token length analysis of MMFineReason. We present the internal domain distribution of CoT lengths (\textit{left}), followed by external comparisons of CoT depth (\textit{mid}) and caption richness (\textit{right}) against prior works. MMFineReason consistently shows higher token counts, indicating greater complexity.}
    \label{fig:token-analysis}
\end{figure}

\subsection{Visual Semantic Analysis}\label{subsec:image}
To quantify the visual diversity of our dataset, we adopt a caption-based classification strategy. By generating detailed descriptions and categorizing the images based on their semantic content, we provide a fine-grained analysis of the visual distribution.

\paragraph{Visual Content Analysis.}
Recent studies~\cite{llava-cot,xia2025visionary} demonstrate that detailed image captioning significantly enhances multimodal reasoning capabilities. Building on this insight, we leverage the powerful \texttt{Qwen3-VL-235B-A22B-Thinking} model to generate structured, high-fidelity captions for the entire MMFineReason dataset. As shown in Figure~\ref{fig:caption-token-comparison} and Table~\ref{tab:token-comparison}, MMFineReason provides significantly denser semantic information, averaging 609 tokens per caption—more than double the length of HoneyBee~\cite{honeybee} (299 tokens). 

Crucially, distinct from HoneyBee which only incorporates captions for a subset of its samples (approx. 58\%), MMFineReason guarantees 100\% caption coverage for all image-question pairs. This comprehensive coverage ensures that every reasoning chain is explicitly supported by fine-grained visual details, providing a more consistent and robust foundation for multimodal learning than baselines with partial visual context.

\paragraph{Visual Topic Diversity.}
Leveraging the generated captions as high-level semantic tags, we taxonomize the dataset into distinct STEM/diagrammatic and natural-image categories; the distribution is detailed in Table~\ref{tab:image_distribution}. The corpus is predominantly composed of STEM and diagrammatic content: geometric diagrams, mathematical plots, and logic puzzles collectively account for $75.3\%$ of the full set. This reflects the MMFineReason's emphasis on symbolic and diagram-centric reasoning. While natural images constitute a smaller fraction, they exhibit significant internal diversity—ranging from urban scenes and documents to astronomical visualizations. This intentional distribution ensures MMFineReason prioritizes fine-grained mathematical reasoning while retaining a complementary natural subset to assess generalization beyond synthetic diagrams. We leave the investigation of optimal mixing ratios between these distributions to future work.

\begin{table}[t]
\caption{Image category statistics by group (STEM vs. Natural). Percentages are normalized within each group.}
\label{tab:image_distribution}
\small
\centering
\resizebox{\textwidth}{!}{
\begin{tabular}{lrr|lrr}
\toprule
\textbf{STEM / Diagrammatic Image} & \textbf{Count} & \textbf{Ratio} & \textbf{Natural Image} & \textbf{Count} & \textbf{Ratio} \\
\midrule
Geometric Diagram & 869{,}959 & 50.02\% & Urban / Street Scene & 5{,}307 & 18.09\% \\
Mathematical Plot / Chart & 332{,}375 & 19.11\% & Indoor / Interior Scene & 4{,}772 & 16.27\% \\
Puzzle / Logic Diagram & 107{,}276 & 6.17\% & Human Portrait / Activity & 3{,}789 & 12.92\% \\
Diagram / Flowchart & 100{,}941 & 5.80\% & Sports / High-Motion Scene & 3{,}659 & 12.47\% \\
Table / Matrix & 76{,}683 & 4.41\% & Document / Text Image & 2{,}900 & 9.89\% \\
Textbook Illustration & 66{,}715 & 3.84\% & Animal / Wildlife Scene & 2{,}125 & 7.24\% \\
Abstract Mathematical Representation & 57{,}569 & 3.31\% & Natural Landscape Scene & 1{,}969 & 6.71\% \\
Spatial Reasoning Scene & 35{,}197 & 2.02\% & Food / Beverage Item & 1{,}835 & 6.26\% \\
Physics / Mechanics Diagram & 34{,}386 & 1.98\% & Vehicle / Machinery Object & 1{,}460 & 4.98\% \\
Biological Structure & 17{,}805 & 1.02\% & Product / Still Life Object & 1{,}125 & 3.84\% \\
Experimental Setup & 10{,}963 & 0.63\% & Artwork / Illustration & 325 & 1.11\% \\
Geological / Earth Science Diagram & 9{,}587 & 0.55\% & Technical / Surveillance / Medical & 68 & 0.23\% \\
Circuit / Network Diagram & 9{,}421 & 0.54\% & {} & {} & {} \\
Astronomy / Space Visualization & 6{,}718 & 0.39\% & {} & {} & {} \\
Molecular / Chemical Diagram & 3{,}735 & 0.21\% & {} & {} & {} \\
\bottomrule
\end{tabular}
}
\end{table}

\subsection{Response Analysis}\label{subsec:response}
Following the generation protocol described in Section~\ref{subsec:distill}, we present a statistical analysis of the resulting generations, specifically characterizing the distribution of domain-specific response lengths and comparing them with existing recent multimodal reasoning datasets, such as OpenMMReasoner~\cite{openmmreasoner} and HoneyBee~\cite{honeybee}.

\paragraph{Reasoning Depth Comparison to Other Datasets.} 
We quantify reasoning depth by analyzing the CoT token length distributions using the Qwen3-VL tokenizer (Figure~\ref{fig:cot-token-comparison} and Table~\ref{tab:token-comparison}). MMFineReason exhibits a substantially more elaborate reasoning process than existing baselines, achieving an average CoT length of 2,910 tokens—approximately $2.7\times$ longer than HoneyBee ($1,063$) and $4.3\times$ longer than OpenMMReasoner ($675$). Notably, the median token count of MMFineReason ($2,038$) is nearly $2.1\times$ that of HoneyBee ($972$) and over $11.3\times$ that of OpenMMReasoner ($180$). This disparity indicates that while baselines often provide concise or superficial rationales, MMFineReason consistently delivers extensive, step-by-step derivations. Furthermore, the extended tail of our distribution (Max: $16,316$) underscores the dataset's capacity to handle highly complex, multi-stage reasoning tasks requiring deep cognitive traversal.

\paragraph{Token Length Distribution of MMFineReason.}
To evaluate the structural quality and domain adaptability of MMFineReason, we analyze the token length distribution and reasoning density across four curated domains. Using the Qwen3-VL tokenizer on the generated responses (\texttt{qwen3vl\_235b\_thinking\_response}), we observe distinct characteristics (visualized in Figure~\ref{fig:cot-token-distribution}):
\begin{itemize}
\item \textbf{Puzzle \& Game:} This domain exhibits the highest average length ($4,810$ tokens), reflecting the most intensive reasoning requirements. The distribution is driven by the necessity for rigorous visual-spatial verification. In tasks such as Raven (avg. $7,745$ tokens) and VisualSphinx (avg. $6,833$ tokens), the model must explicitly hypothesize rules and verify candidate options sequentially, resulting in dense ``System-2'' reasoning traces that significantly exceed those of other domains.

\item \textbf{Mathematics:} The Mathematics domain demonstrates high information density with an average length of $2,950$ tokens. Distinguished by exceptional symbolic rigor, this category exhibits a density of LaTeX markers more than double that of scientific tasks. The structure is distinct from natural language tasks, heavily populated with step-by-step symbolic derivations (e.g., Euclid30K, Geo170k) essential for precise calculation.

\item \textbf{Science:} Averaging $2,305$ tokens across $245$k samples, this domain bridges abstract logic and real-world context. The data reflects a dual process: the model must ground visual entities (Perception) before applying domain-specific knowledge (Causal Inference). The result is a balanced reasoning structure combining substantial textual explanation with moderate symbolic usage.

\item \textbf{General/OCR:} Serving as a regularization baseline, this category remains concise (avg. $1,262$ tokens). Primarily derived from LLaVA-CoT, these samples prioritize direct visual grounding over complex logical deduction. This preserves the model's ``System-1'' perception capabilities, mitigating ``reasoning hallucination''—the tendency to over-generate complex rationales for simple perceptual tasks.
\end{itemize}

\section{Experiments} 
In this section, we empirically validate the effectiveness of our proposed reasoning datasets, MMFineReason. We begin by detailing the experimental setup in Section~\ref{subsec:setup}, followed by a presentation of the main results in Section~\ref{subsec:results}. Section~\ref{subsec:rl} analyzes the training dynamics of the Reinforcement Learning (RL) stage. We then examine the impact of our data filtering strategy via ablation studies in Section~\ref{subsec:ablation}. Finally, we investigate the fine-grained contributions of individual sub-datasets in Section~\ref{subsec:individual}.

\subsection{Experimental Setup}\label{subsec:setup}
\paragraph{Training Details.}
We use LLaMA-Factory~\cite{llamafactory} and VeRL~\cite{verl} as our SFT and RL training frameworks, respectively. The training configurations are summarized in Table~\ref{tab:sft_params} and Table~\ref{tab:rl_params}. We train the Qwen3-VL-2B/4B/8B-Instruct models under the same experimental setup.

\vspace{5pt}
\noindent\textbf{Supervised Fine-Tuning (SFT).}
As shown in Table~\ref{tab:sft_params}, we optimize the model using AdamW with a learning rate of $1\text{e-}5$ and a cosine decay scheduler. To maximize training throughput and reduce memory fragmentation, we utilize liger kernel and enable sequence packing with a length of 32,768. The input images are resized to a resolution of $768 \times 768$ during this stage to balance efficiency and performance. We train the models for 3 epochs with a global batch size of 32.

\vspace{5pt}
\noindent\textbf{Reinforcement Learning (RL).}
For the RL stage, we adopt the GSPO algorithm~\cite{zheng2025groupsequencepolicyoptimization} to enhance the reasoning capabilities of the model. As detailed in Table~\ref{tab:rl_params}, we set the learning rate to $1\text{e-}6$ with a constant scheduler to ensure stable convergence. A key component of our setup is the generation of $G=16$ rollouts for each prompt to estimate the group-dependent baseline, which reduces the variance of the gradient estimator. The training spans 300 steps with a batch size of 256.

\vspace{5pt}
\noindent\textbf{Evaluation Setup.}
We evaluate our models on VLMEvalKit~\cite{vlmevalkit}. To strictly assess the model's reasoning reliability, we employ greedy decoding (Temperature $= 0$) for all benchmarks. Notably, as shown in Table~\ref{tab:eval_params}, we increase the maximum image resolution to $2048 \times 2048$ during inference. More evaluation details are in Appendix~\ref{apx:eval}.

\paragraph{Baselines.} Our baselines fall into three main categories: (1) \textbf{Closed-source VLMs}, including Gemini-2.5-Flash~\cite{gemini2.5} and GPT5-mini High~\cite{gpt5}; (2) \textbf{Open-weight VLMs}, specifically Qwen3-VL-8B-Thinking, Qwen3-VL-30B-A3B-Thinking and Qwen3-VL-32B-Thinking; and (3) \textbf{Open-source VLMs}. For MMR1~\cite{mmr1} and HoneyBee~\cite{honeybee}, we fine-tune Qwen3-VL-8B-Instruct on their respective SFT datasets to ensure a fair comparison. For OMR-7B~\cite{openmmreasoner}, we directly report the RL results from the original paper and our evaluation using the officially released checkpoints. All models are evaluated in thinking mode to ensure consistent comparison.

\paragraph{Benchmarks \& Evaluation.}
To ensure a comprehensive assessment, we evaluate our model across a diverse suite of benchmarks spanning three key domains:
\begin{itemize}
    \item \textbf{STEM \& Puzzle}: 
    MMMU\textsubscript{val}~\cite{mmmu}, MathVista\textsubscript{mini}~\cite{mathvista}, MathVision\textsubscript{test}~\cite{mathvision}, MathVerse\textsubscript{mini}~\cite{mathverse}, 
    Dynamath~\cite{dynamath}, LogicVista~\cite{logicvista}, VisuLogic~\cite{visulogic}, ScienceQA~\cite{scienceqa}.
    
    \item \textbf{General VQA}: 
    RealWorldQA~\cite{realworldqa}, 
    , MMBench-EN~\cite{mmbench}, MMStar~\cite{mmstar}.
    
    \item \textbf{Document Understanding}: 
    AI2D~\cite{ai2d},
    CharXiv\textsubscript{reas.}~\cite{charxiv}, 
    CharXiv\textsubscript{desc.}~\cite{charxiv}.
\end{itemize}

\subsection{Main Results}\label{subsec:results}

\definecolor{highlightcolor}{RGB}{255, 224, 204}
\definecolor{TableBlue}{RGB}{235, 245, 255}
\definecolor{IceBlue}{RGB}{220, 240, 255}
\begin{table*}[h!]
\centering
\caption{Comparison of MMFineReason (MFR) models with state-of-the-art models on various multimodal benchmarks.}
\label{tab:main_results}
\setlength{\tabcolsep}{4pt}
\resizebox{\textwidth}{!}{%
\begin{tabular}{l c c c c c c c c >{\columncolor{IceBlue}}c >{\columncolor{IceBlue}}c >{\columncolor{IceBlue}}c}
\toprule
\multirow{2}{*}{\textbf{Benchmarks}} & \multicolumn{2}{c}{\textbf{Closed-source VLMs}} & \multicolumn{3}{c}{\textbf{Open-weight VLMs}} & \multicolumn{3}{c}{\textbf{Open-source VLMs}} & \multicolumn{3}{c}{\textbf{Ours}} \\
\cmidrule(lr){2-3} \cmidrule(lr){4-6} \cmidrule(lr){7-9} \cmidrule(lr){10-12}
 & \makecell{\textbf{Gemini-2.5}\\\textbf{Flash}} & \makecell{\textbf{GPT5}\\\textbf{mini}} & \makecell{\textbf{Qwen3-VL}\\\textbf{8B}} & 
 \makecell{\textbf{Qwen3-VL}\\\textbf{30B-A3B}} & \makecell{\textbf{Qwen3-VL}\\\textbf{32B}} & \makecell{\textbf{MMR1}\\\textbf{8B}} & \makecell{\textbf{HoneyBee}\\\textbf{8B}} & \makecell{\textbf{OMR}\\\textbf{7B}} &
 \makecell{\textbf{MFR}\\\textbf{2B}} &
 \makecell{\textbf{MFR}\\\textbf{4B}} &
 \makecell{\textbf{MFR}\\\textbf{8B}} \\
\midrule
\textbf{MMMU}$_{\text{val}}$           & 77.7  & 79.0 & 74.1 & 76.0 & 78.1 & 62.8 & 63.1 & 57.8 & 54.8 & 69.6 & 71.3 \\
\textbf{MathVista}$_{\text{mini}}$ & 79.4 & 79.1 & 81.4 & 81.9 & 85.9 & 75.3 & 71.9 & 79.5 & 74.6 & 82.2 & 81.7 \\
\textbf{MathVision}$_{\text{test}}$    & 64.3 & 71.9 & 62.7 & 65.7 & 70.2 & 48.4 & 37.4 & 43.6 & 45.3 & 61.3 & 67.1 \\
\textbf{MathVerse}$_{\text{mini}}$ & 77.7 & 78.8 & 77.7 & 79.6 & 82.6 & 67.3 & 60.9 & 63.8 & 69.2 & 78.7 & 81.5 \\
\textbf{DynaMath}$_{\text{test}}$      & 75.9 & 81.4 & 73.2 & 80.1 & 82.0 & 73.6 & 69.4 & 69.1  & 71.4 & 80.6 & 83.4 \\
\textbf{LogicVista}$_{\text{test}}$    & 67.3 & 71.4 & 65.1 & 65.8 & 70.9 & 54.6 & 47.8 & 50.0 & 53.8 & 67.6 & 68.5 \\
\textbf{VisuLogic}$_{\text{test}}$     & 31.0 & 27.2 & 27.5 & 26.6 & 32.4 & 25.4 & 25.9 & 24.4  & 28.3 & 29.8 & 30.5 \\
\textbf{ScienceQA}   & 97.1 & 96.9 & 94.8 & 96.4 & 97.2 & 95.4 & 95.2 & 96.8  & 94.4 & 95.8 & 97.5 \\
\midrule
\textbf{RWQA}$_{\text{test}}$          & 76.0 & 79.0 & 73.5 & 77.4 & 78.4 & 71.0 & 70.5 & 69.4 & 68.2 & 74.9 & 75.6 \\
\textbf{MMBench-EN}          & 87.0 & 86.6 & 85.3 & 87.0 & 89.5 & 86.9 & 87.4 & 85.9  & 84.5 & 88.7 & 88.9\\
\textbf{MMStar}$_{\text{test}}$          & 76.5 & 74.1 & 75.3 & 75.5 & 79.4 & 69.3 & 73.3 & 69.0   & 67.7 & 72.8 & 75.2\\
\midrule
\textbf{AI2D}$_{\text{test}}$          & 88.7 & 88.2 & 84.9 & 86.9 & 88.9 & 83.4 & 86.0  & 85.0 & 82.5 & 86.5 & 87.9\\
\textbf{CharXiv}$_{\text{reas.}}$      & 61.7 & 68.6 & 53.0 & 56.6 & 65.2 & 48.8 & 47.4 & 46.1 & 45.4 & 58.1 & 60.0 \\
\textbf{CharXiv}$_{\text{desc.}}$      & 90.1 & 89.4 & 85.9 & 86.9 & 90.2 & 81.5 & 75.8 & 73.5 & 74.3 & 87.7 & 90.8     \\
\midrule
\textbf{Avg}  & 75.0 & 76.5 & 72.5 & 74.5 & 77.9 & 67.4 & 65.1 & 65.3  & 65.3 & 73.9 & 75.7\\
\bottomrule
\end{tabular}%
}
\end{table*}

Table~\ref{tab:main_results} presents a comprehensive comparison of our MMFineReason models (MFR) against SOTA open-source models and proprietary vision-language models. Our models establish new SOTA results for their size class. Our MFR-2B model already approaches existing open-source 8B models, the MFR-4B model surpasses Qwen3-VL-8B-Thinking, and our MFR-8B model even outperforms Qwen3-VL-30B-A3B-Thinking while approaching the performance of Qwen3-VL-32B-Thinking, demonstrating remarkable parameter efficiency.

\noindent\textbf{Dominance in Mathematical \& Logical Reasoning.} MFR models exhibit substantial improvements over competing methods. Our MFR-8B surpasses the same-sized Qwen3-VL-8B-Thinking by a large margin and outperforms the significantly larger Qwen3-VL-30B-A3B-Thinking on nearly all mathematical benchmarks. On DynaMath, MFR-8B achieves 83.4\%, outperforming Qwen3-VL-32B-Thinking at 82.0\% and Qwen3-VL-30B-A3B-Thinking at 76.7\%. On MathVerse, MFR-8B reaches 81.5\%, approaching Qwen3-VL-32B-Thinking at 82.6\% while surpassing Qwen3-VL-30B-A3B-Thinking at 79.6\%. 

\noindent\textbf{Strong Generalization Across Domains.}
A surprising and notable observation is that our MFR models exhibit strong generalization ability, maintaining competitive performance on both general understanding and chart reasoning benchmarks. Specifically, on RWQA, MFR-8B achieves 75.6\%, improving over open-source baselines such as MMR1-8B at 71.0\% and HoneyBee-8B at 70.5\%. On CharXiv$_{\text{desc.}}$, MFR-8B achieves 89.9\%, approaching Qwen3-VL-32B-Thinking at 90.2\% and closed-source models. It is worth noting that our training data contains minimal chart or real-world related samples, yet the enhanced reasoning capabilities generalize effectively to these general domains.

\noindent\textbf{Superior Data Efficiency vs. Open-source Baselines.}
A key finding is the significant performance gap between MFR and other open-source baselines. On MathVision, MFR-8B achieves 67.1\%, outperforming HoneyBee-8B at 37.4\% and OMR-7B at 36.6\% by over 30 absolute points. On MathVerse, MFR-8B at 81.5\% surpasses MMR1-8B at 67.3\% and HoneyBee-8B at 60.9\% by a large margin. These results demonstrate that the quality of reasoning chains in MFR is far superior to scale-focused strategies.

\begin{table*}[h!]
\centering
\caption{Main results of different model scales across various multimodal benchmarks. We compare our MMFineReason (MFR) models against the base Qwen3-VL variants. \textbf{"Inst."} and \textbf{"Think."} denote Qwen3-VL-Instruct and Qwen3-VL-Thinking, respectively; \textbf{"SFT"} and \textbf{"RL"} refer to our MFR-SFT and MFR-Thinking models.}
\label{tab:split_results}
\setlength{\tabcolsep}{7pt}
\resizebox{\textwidth}{!}{%
\begin{tabular}{l c c >{\columncolor{IceBlue}}c >{\columncolor{IceBlue}}c c c >{\columncolor{IceBlue}}c >{\columncolor{IceBlue}}c c c >{\columncolor{IceBlue}}c >{\columncolor{IceBlue}}c}
\toprule
\multirow{2}{*}{\textbf{Benchmarks}} & \multicolumn{4}{c}{\textbf{2B Models}} & \multicolumn{4}{c}{\textbf{4B Models}} & \multicolumn{4}{c}{\textbf{8B Models}} \\
\cmidrule(lr){2-5} \cmidrule(lr){6-9} \cmidrule(lr){10-13}
 & \textbf{Inst.} & \textbf{Think.} & \textbf{SFT} & \textbf{RL} & \textbf{Inst.} & \textbf{Think.} & \textbf{SFT} & \textbf{RL} & \textbf{Inst.} & \textbf{Think.} & \textbf{SFT} & \textbf{RL} \\
\midrule
\textbf{MMMU}$_{\text{val}}$       & 53.4 & 61.4 & 54.6 & 54.8 & 67.4 & 70.8 & 69.3 & 69.6 & 69.6 & 74.1 & 71.3 & 71.3 \\
\textbf{MathVista}$_{\text{mini}}$ & 61.3 & 73.6 & 73.3 & 74.6 & 73.7 & 79.5 & 80.1 & 82.2 & 77.2 & 81.4 & 81.2 & 81.7 \\
\textbf{MathVision}$_{\text{test}}$ & 31.6 & 45.9 & 40.9 & 45.3 & 51.6 & 60.0 & 62.4 & 61.3 & 53.9 & 62.7 & 67.6 & 67.1 \\
\textbf{MathVerse}$_{\text{mini}}$ & 52.1 & 66.9 & 70.4 & 69.2 & 46.8 & 75.2 & 78.4 & 78.7 & 62.1 & 77.7 & 82.2 & 81.5 \\
\textbf{DynaMath}$_{\text{test}}$  & 54.2 & 66.7 & 68.7 & 71.4 & 65.3 & 74.4 & 79.9 & 80.6 & 67.7 & 73.2 & 82.6 & 83.4 \\
\textbf{LogicVista}$_{\text{test}}$ & 35.8 & 50.0 & 52.8 & 53.8 & 53.2 & 61.1 & 66.7 & 67.6 & 55.3 & 65.1 & 68.7 & 68.5 \\
\textbf{VisuLogic}$_{\text{test}}$  & 11.5 & 25.4 & 24.7 & 28.3 & 19.0 & 30.2 & 27.8 & 29.8 & 22.5 & 27.5 & 29.9 & 30.5 \\
\textbf{ScienceQA}                 & 87.4 & 88.0 & 92.3 & 94.4 & 88.0 & 94.1 & 96.6 & 95.8 & 95.4 & 94.8 & 95.4 & 97.5 \\
\midrule
\textbf{RWQA}$_{\text{test}}$      & 63.9 & 69.5 & 67.9 & 68.2 & 70.9 & 73.2 & 71.5 & 74.9 & 71.5 & 73.5 & 74.1 & 75.6 \\
\textbf{MMBench-EN}                & 78.4 & 79.9 & 83.2 & 84.5 & 83.9 & 84.6 & 88.7 & 88.7 & 84.5 & 85.3 & 87.8 & 88.9 \\
\textbf{MMStar}$_{\text{test}}$    & 58.3 & 68.1 & 63.6 & 67.7 & 69.8 & 73.2 & 73.0 & 72.8 & 70.9 & 75.3 & 74.8 & 75.2 \\
\midrule
\textbf{AI2D}$_{\text{test}}$      & 76.9 & 80.4 & 78.5 & 82.5 & 84.1 & 84.9 & 86.1 & 86.5 & 85.7 & 84.9 & 86.5 & 87.9 \\
\textbf{CharXiv}$_{\text{reas.}}$  & 26.8 & 37.1 & 39.0 & 45.4 & 39.7 & 50.3 & 55.9 & 58.1 & 46.4 & 53.0 & 58.4 & 60.0 \\
\textbf{CharXiv}$_{\text{desc.}}$  & 62.3 & 70.1 & 74.1 & 74.3 & 76.2 & 83.9 & 87.7 & 87.7 & 83.0 & 85.9 & 89.9 & 90.8 \\
\midrule
\textbf{Avg}                       & 53.9 & 63.1 & 63.1 & 65.3 & 63.5 & 71.1 & 73.2 & 73.9 & 67.6 & 72.5 & 75.0 & 75.7 \\
\bottomrule
\end{tabular}%
}
\end{table*}

\subsection{Effectiveness of Different Training Stages}\label{subsec:rl}

Table~\ref{tab:split_results} presents the results of SFT and RL training across different model scales. We observe that each training stage contributes to the final performance.

\noindent\textbf{SFT Drives Major Gains in Reasoning.}
For mathematical and logical reasoning, the largest performance improvements come from SFT. Compared to Qwen3-VL-Instruct, MFR-SFT achieves substantial gains across all model scales. For the 8B model, SFT improves MathVision from 53.90\% to 67.56\% and LogicVista from 55.30\% to 68.68\%. Similar trends are observed at smaller scales: the 2B model gains +3.5\% on MathVerse and +2.8\% on LogicVista after SFT.

\noindent\textbf{RL Enhances Generalization.}
We find that RL training significantly improves generalization on general understanding and chart benchmarks. For the 2B model, RL improves AI2D from 78.47\% to 82.51\% and CharXiv$_{\text{reas.}}$ from 38.96\% to 45.38\%. For the 8B model, RL brings consistent gains on RWQA, SciQA, and CharXiv$_{\text{desc.}}$, demonstrating that RL effectively enhances the model's ability to generalize beyond the reasoning-focused training distribution.

\noindent\textbf{RL Shows Variance on Math Benchmarks.}
However, we also observe that RL exhibits some variance on mathematical benchmarks. While RL improves DynaMath across all scales, it causes slight drops on MathVision for 4B and 8B models. We hypothesize that since the model has already learned most patterns during SFT, further RL gains require more diverse or challenging data. Exploring more effective RL data strategies remains an important direction for future work.

\begin{figure}[t!]
    \centering
    \includegraphics[width=0.8\linewidth]{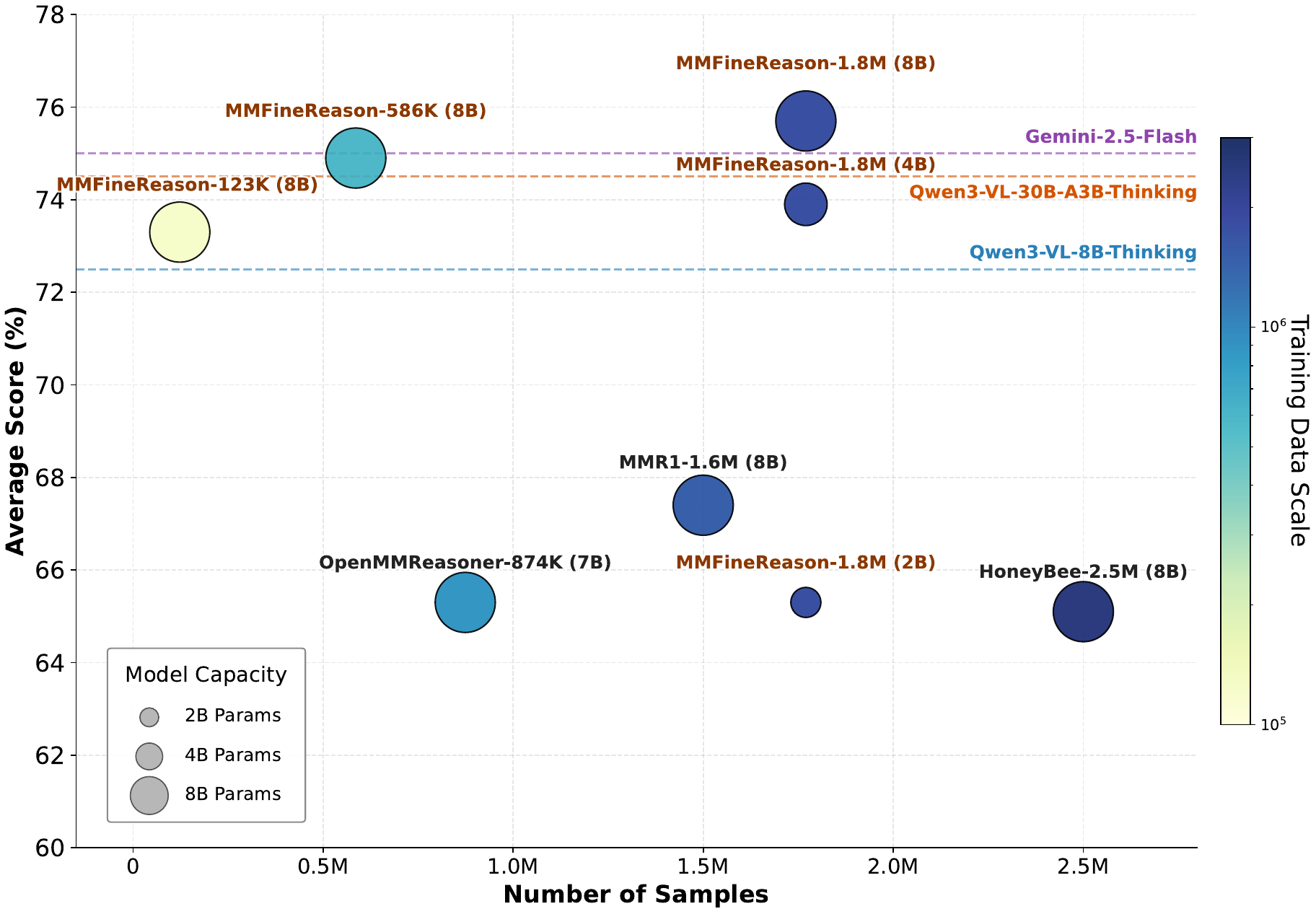}
    \caption{Performance comparison of MMFineReason against existing open-source datasets across different data scales and model sizes. Bubble size indicates model parameter count (2B, 4B, 8B), while color intensity represents training data volume. Dashed lines denote the performance of strong baselines. MMFineReason demonstrates superior data efficiency, achieving higher accuracy with significantly fewer samples and smaller model parameters compared to MMR1 and HoneyBee.}
    \label{fig:compare_data}
\end{figure}

\subsection{Scaling Frontiers and Data Efficiency}
\label{subsec:scaling_analysis}

In this section, we evaluate the effectiveness of our proposed data strategy by benchmarking MMFineReason against existing SOTA open-source datasets and models. As illustrated in Figure~\ref{fig:compare_data}, our approach demonstrates superior scaling properties across data volume, data quality, and model capacity.

\paragraph{Superior Data Quality and Peak Performance.}
We first examine the impact of data quality by comparing models trained on full datasets under the same parameter scale (8B). MMFineReason-1.8M achieves a peak score of 75.7, establishing a substantial lead over existing open-source datasets such as MMR1-1.6M (67.4) and HoneyBee-2.5M (65.1). Despite utilizing a smaller sample size than HoneyBee (1.8M vs. 2.5M), our model achieves a performance gain of \textbf{+10.6} points. This significant gap highlights that the fine-grained reasoning logic in our dataset provides much denser supervision signals than standard caption-based or coarse-reasoning datasets.

\paragraph{Extreme Data Efficiency.}
Beyond peak performance, MMFineReason exhibits remarkable efficiency in low-data regimes. The most striking finding is that our minimal subset, MMFineReason-123K, already achieves a score of \textbf{73.3}. This result significantly outperforms models trained on the full HoneyBee (2.5M) and MMR1 (1.6M) datasets. By utilizing only $\sim$\textbf{5\%} of the data volume of comparable benchmarks, MMFineReason matches or exceeds their performance. This "cross-over" effect suggests that rigorous filtering and high-quality rationale construction effectively eliminate the redundancy found in large-scale datasets, allowing for faster convergence with a fraction of the computational budget.

\paragraph{SOTA Comparison and Parameter Efficiency.}
Finally, we benchmark our models against leading open-weights and commercial baselines (represented by dashed lines in Figure~\ref{fig:compare_data}). Our high-quality data enables smaller models to punch above their weight class. Specifically, our 4B model achieves a score of 73.9, surpassing the widely-used Qwen3-VL-8B-Thinking (72.5), which demonstrates that superior data quality can effectively compensate for a $2\times$ reduction in model parameters. Scaling up to 8B parameters, our model establishes a new state-of-the-art for its size class with a score of 75.7. It not only outperforms the significantly larger Qwen3-VL-30B-A3B-Thinking (74.5) but also exceeds the performance of the commercial baseline Gemini-2.5-Flash (75.0). These results validate that MMFineReason enables efficient open-weights models to compete directly with, and even surpass, proprietary and significantly larger foundation models.

\subsection{Ablation Studies}\label{subsec:ablation}

To validate the effectiveness of our proposed data strategy and training settings, we conduct two sets of ablation studies to investigate the impact of caption augmentation strategies and input resolution on model performance.

\begin{figure}[t!]
    \centering
    \includegraphics[width=1\linewidth]{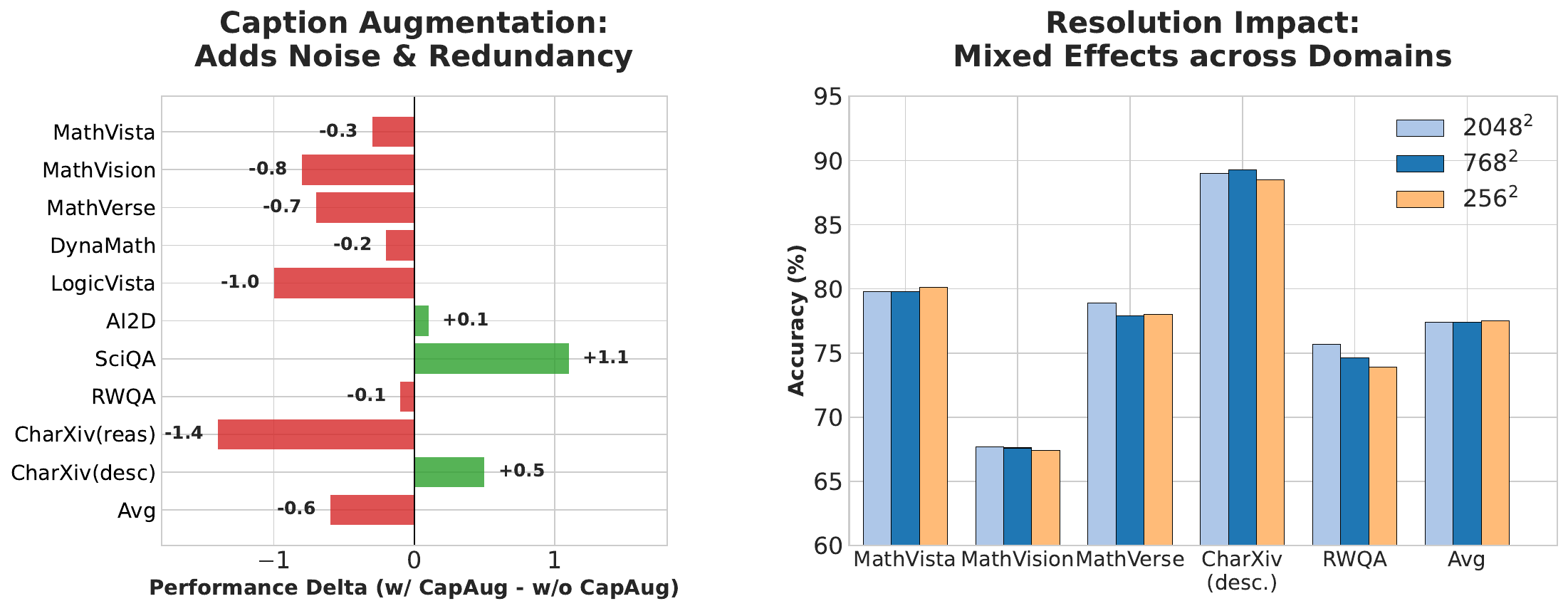}
    \caption{Ablation studies on data and training strategies. 
    \textit{Left:} caption augmentation brings marginal or negative gains on STEM benchmarks, likely redundant with visual cues in long CoT. 
    \textit{Right:} ultra-high resolution ($2048^2$) shows diminishing returns over $768^2$/$256^2$ for geometry and chart tasks, though still beneficial for natural images (RWQA).}
    \label{fig:ablation}
\end{figure}

\vspace{5pt}
\paragraph{Trade-off of Caption Augmentation.}
Figure~\ref{fig:ablation} (left) presents the effects of introducing caption augmentation (CapAug), specifically by prepending image captions enclosed within \texttt{<caption>...</caption>} tags to the response. The experiments reveal a clear trade-off in model performance. Specifically, CapAug slightly boosts logical reasoning capabilities, with LogicVista improving by 1.1\%. This can likely be attributed to captions forcing the model to parse the layout and complete information of images, facilitating a better understanding of structure and logical relationships. However, on STEM visual reasoning tasks like MathVista and MathVision, it does not bring significant performance gains. This is likely because the Long CoT already contains sufficient visual information required for reasoning, rendering the caption information redundant. Consequently, we do not adopt CapAug in our final version.

\paragraph{Scaling Effect of Input Resolution.}
The analysis of input resolution (Max Pixels) in Figure~\ref{fig:ablation} (right) shows that simply increasing resolution does not always yield performance gains. Surprisingly, ultra-high resolution ($2048^2$) does not outperform medium-to-low resolution settings on multiple benchmarks. For example, on MathVista and CharXiv, the performance of $768^2$ or even $256^2$ is superior to that of $2048^2$. Although $2048^2$ achieves a slight advantage on MathVision, the overall benefit is limited considering the significantly increased computational cost. This phenomenon suggests that: (1) the core features of many current benchmark problems (especially geometry and charts) do not rely on pixel-level ultra-high definition; and (2) excessive resolution results in overly long visual token sequences, introducing redundancy and increasing the difficulty for the attention mechanism to capture key features. 
However, a notable exception is observed on RealWorldQA, which focuses on general natural images. On this benchmark, scaling up to $2048^2$ yields stable performance gains. This is likely because real-world scenarios often contain fine-grained details, small objects, or dense text embedded in complex backgrounds, which necessitate higher pixel density for precise recognition. In contrast to the sparse and structural nature of diagrams, natural images require high resolution to resolve these subtle visual cues effectively. Balancing performance across domains and computational efficiency, we adopt $768^2$ as our default setting.

\subsection{Distilled Sub-Dataset Analysis}\label{subsec:individual}
\begin{figure}
    \centering
    \includegraphics[width=1\linewidth]{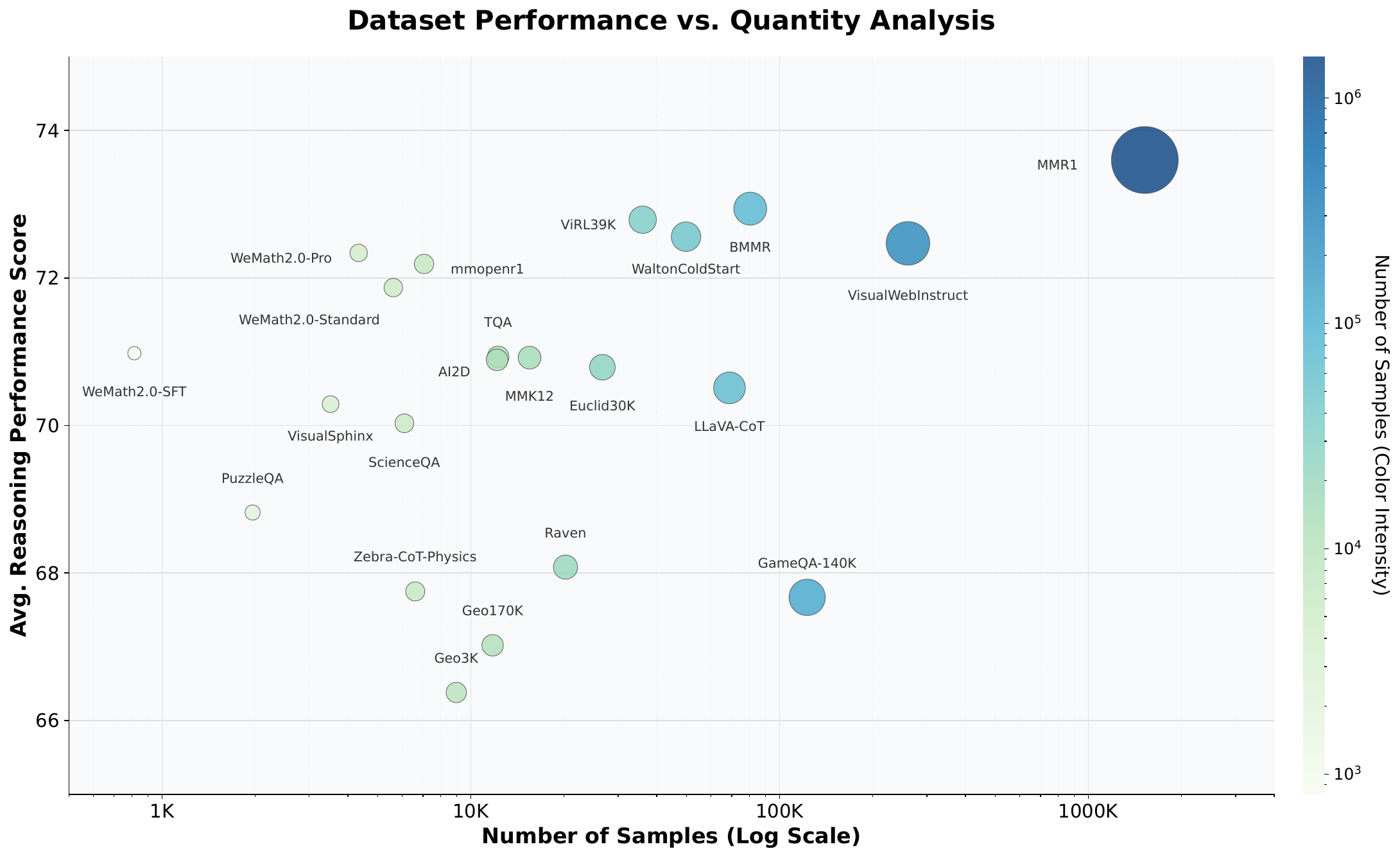}
    \caption{Performance landscape of distilled sub-datasets.}
    \label{fig:subset_exp}
\end{figure}
The experimental results in Figure~\ref{fig:subset_exp} characterize how training data properties—specifically sample size and domain composition—shape downstream STEM reasoning performance. Across all subsets, we observe the following key findings.

\paragraph{Diminishing Returns of Scale and the Pareto Frontier.}By juxtaposing MMR1 ($N \approx 1.5\text{M}$) with ViRL39K ($N \approx 39\text{K}$), we observe a significant non-linear relationship between data volume and reasoning performance. While MMR1 establishes the upper bound with $73.60\%$ accuracy, ViRL39K achieves $72.79\%$ accuracy—retaining $98.9\%$ of the relative performance using only $2.4\%$ of the data volume. This constructs a new Pareto frontier, indicating that systematically cleaned, reformatted, and verified data (as seen in ViRL39K) can drastically reduce training costs. It further suggests that blind data scaling encounters severe diminishing returns beyond a certain threshold in multimodal reasoning tasks.

\paragraph{Latent Capability Activation via High-Density Instruction.}The most significant outlier in our scaling analysis is WeMath2.0-SFT. Despite comprising a negligible $0.05\%$ of the total data volume ($N=814$), it achieves a reasoning accuracy of $70.98\%$, effectively matching the performance of datasets three orders of magnitude larger, such as MMR1 ($73.60\%$). This disproportionate efficiency validates the ``Knowledge-Oriented Chain-of-Thought'' (KO-CoT) paradigm. We hypothesize that large-scale pre-training endows models with latent domain knowledge, but often leaves them without the specific reasoning syntax required for complex problem-solving. WeMath2.0-SFT functions not as a knowledge source, but as a high-efficiency catalyst, aligning the model's internal representations with structured reasoning paths. This suggests that for foundation models, a small set of high-quality, reasoning-dense instructions is sufficient to trigger capabilities that otherwise remain dormant in massive, noisy corpora.

\paragraph{Not all ``hard'' reasoning transfers: puzzle/game datasets lag.}
Puzzle/game-centric datasets are consistently weaker: GameQA-140K (122.9K, 67.67), Raven (20.3K, 68.08), and PuzzleQA (2.0K, 68.82). Despite their rigorous generation/verification (e.g., engine-derived annotations in GameQA-140K), these tasks emphasize planning-like search, symbolic state transitions, and abstract relational rules that may be \textbf{mismatched} with the dominant evaluation distribution (often math/science QA). Another plausible factor is that many puzzle/game solutions resemble program-execution traces; if the target model is not explicitly trained to internalize algorithmic state updates, these examples contribute less to general multimodal QA performance.

\paragraph{Geometry-only corpora underperform, indicating narrow visual grammars.}
Geo3K (9.0K, 66.38) and Geo170K (11.8K, 67.02) are among the lowest performers, despite being in-domain for diagram reasoning. This highlights a critical nuance: geometry datasets can be \textbf{structurally narrow} (limited diagram styles, repetitive constructions, constrained linguistic patterns), which reduces their marginal utility for broad reasoning. Euclid30K performs substantially better (26.7K, 70.79), supporting the hypothesis that \textbf{formalized reasoning structure} and better-designed problem diversity (plane/solid geometry, proof-like steps) are more important than simply adding more geometry instances.

\paragraph{Impact of Disciplinary Breadth on Generalization.}
Comparing GameQA-140K ($67.67\%$) and BMMR ($72.94\%$) highlights the importance of disciplinary diversity in Chain-of-Thought (CoT) data. Although GameQA offers a larger volume of samples ($140\text{K}$), its scope is restricted to closed-world game logic. BMMR, despite being smaller ($80\text{K}$), spans over 300 academic disciplines. This breadth allows the model to internalize a more generalized reasoning structure, suggesting that for general-purpose vision-language models, diversity in the reasoning domain is a stronger driver of performance than the depth of any single task type.

\section{Conclusion}
\label{section:conclusion}
In this work, we view multimodal reasoning primarily as a data-centric problem rather than a purely model-centric one. MMFineReason shows that strong multimodal reasoning capabilities can be systematically induced through structured data design, without relying on excessive model scaling or proprietary supervision.
Beyond the dataset itself, this study establishes a scalable framework for reasoning data engineering, covering reasoning supervision, difficulty-aware structuring, and data composition. Results across multiple model scales demonstrate that well-curated reasoning data can deliver substantial performance gains and parameter efficiency, enabling smaller open models to compete with much larger reasoning-oriented systems.
Moreover, our findings indicate that reasoning-oriented supervision functions as a general capability amplifier, benefiting both complex reasoning tasks and broader multimodal understanding. MMFineReason therefore serves as both a benchmark resource and a reproducible methodology for building open, efficient multimodal reasoning systems.

\clearpage
\newpage
\bibliographystyle{plainnat}
\setcitestyle{numbers}
\bibliography{paper}

\clearpage
\newpage
\beginappendix

\section{Data Curation Details}
\label{sec:subset}
\subsection{Statistics}

\begin{table}[htbp]
    \centering
    \caption{Dataset composition of MMFineReason. Subsets marked with $\dagger$ is inherited from FineVision~\cite{finevision}.}
    \label{tab:data_stats_split}
    \resizebox{\textwidth}{!}{%
    \begin{tabular}{llrr|llrr}
        \toprule
        \textbf{Subset Name} & \textbf{Category} & \textbf{Samples} & \textbf{Tokens} & \textbf{Subset Name} & \textbf{Category} & \textbf{Samples} & \textbf{Tokens} \\
        \midrule
        VisualWebInstruct$^\dagger$~\cite{finevision} & Science & 260,556 & 696,366,646 & 
        AI2D$^\dagger$~\cite{finevision}              & Science & 12,167  & 29,657,396 \\
        
        MMR1~\cite{mmr1}                  & Math    & 1,524,033 & 5,924,286,949 &
        Geo170k(qa)$^\dagger$~\cite{finevision}       & Math    & 11,771  & 25,606,103 \\
        
        GameQA-140K~\cite{gameqa}                     & Puzzle  & 122,868 & 751,735,082 & 
        Geometry3k$^\dagger$~\cite{finevision}        & Math    & 8,977   & 16,434,149 \\
        
        BMMR~\cite{bmmr}                              & Science & 80,366  & 380,074,912 & 
        mmopenr1-8k~\cite{mmopenr1}                   & Math    & 7,057   & 27,520,245 \\
        
        LLaVA-CoT~\cite{llava-cot}                    & General & 68,838  & 132,188,410 & 
        Zebra-CoT-Physics~\cite{zebracot}             & Science & 6,610   & 31,020,152 \\
        
        WaltonColdStart~\cite{walton}                 & Math    & 49,786  & 146,912,078 & 
        ScienceQA$^\dagger$~\cite{finevision}         & Science & 6,095   & 9,681,856 \\
        
        ViRL39K~\cite{virl}                           & Math    & 36,034  & 129,740,324 & 
        WeMath2-Standard~\cite{wemath}                & Math    & 5,613   & 16,599,703 \\
        
        Euclid30K~\cite{euclid}                       & Math    & 26,690  & 124,091,304 & 
        WeMath2-Pro~\cite{wemath}                     & Math    & 4,334   & 18,588,338 \\
        
        Raven$^\dagger$~\cite{finevision}             & Puzzle  & 20,271  & 192,077,480 & 
        VisualSphinx~\cite{visualsphinx}              & Puzzle  & 3,516   & 28,666,407 \\
        
        MMK12~\cite{mmk12}                            & Science & 15,505  & 66,234,373  & 
        PuzzleVQA~\cite{puzzlevqa}                    & Puzzle  & 1,966   & 7,581,908 \\
        
        TQA$^\dagger$~\cite{finevision}               & Science & 12,263  & 26,319,379  & 
        WeMath2-SFT~\cite{wemath}                     & Math    & 814     & 4,040,475 \\
        \midrule
        \multicolumn{8}{c}{\textbf{Total Samples:\quad 2,286,130 ,\quad Total Tokens:\quad 8,785,423,669 }} \\
        \bottomrule
    \end{tabular}%
    }
\end{table}
\subsection{Inclusion and Exclusion Criteria}
We prioritize using pre-filtered subsets when higher-quality versions of a dataset are available---for example, the FineVision-filtered VisualWebInstruct subset and the OpenMMReasoner-filtered MMR1 subset.

We exclude the following categories of data from our collection:
\begin{itemize}
    \item \textbf{Multi-image samples:} Data where each instance contains more than one image.
    \item \textbf{Overly simple task data:} e.g., CLEVR family and geo170k (align), which provide limited reasoning complexity.
    \item \textbf{Highly specialized imaging domains:} e.g., PathVQA, which focuses on medical imagery.
    \item \textbf{Multilingual datasets featuring minor languages within images:} e.g., EXAMS-V.
\end{itemize}

\subsection{Filteration Details}
\label{apx:filter}

\begin{table}[htbp]
    \centering
    \caption{\textbf{Data Cleaning Statistics by Dataset.} "Filt. (Len)" refers to length-based filtering, and "Filt. (Tem)" refers to template validation errors.}
    \label{tab:data_cleaning_stats}
    \resizebox{\textwidth}{!}{%
    \begin{tabular}{lrrrrc}
        \toprule
        \textbf{Dataset} & \textbf{Original} & \textbf{Filt. (Len)} & \textbf{Filt. (Tem)} & \textbf{Remain} & \textbf{Rate (\%)} \\
        \midrule
        BMMR & 84,252 & 67 & 3,819 & 80,366 & 95.39 \\
        Euclid30K & 27,021 & 3 & 328 & 26,690 & 98.78 \\
        FineVision-ai2d\_merged & 12,180 & 13 & 0 & 12,167 & 99.89 \\
        FineVision-geo170k(qa) & 12,101 & 56 & 274 & 11,771 & 97.27 \\
        FineVision-geometry3k(mathv360k) & 9,716 & 271 & 468 & 8,977 & 92.39 \\
        FineVision-raven & 20,411 & 14 & 126 & 20,271 & 99.31 \\
        FineVision-scienceqa & 6,112 & 4 & 13 & 6,095 & 99.72 \\
        FineVision-tqa & 12,560 & 5 & 292 & 12,263 & 97.64 \\
        FineVision-visualwebinstruct(filt) & 261,007 & 449 & 2 & 260,556 & 99.83 \\
        GameQA-140K & 123,579 & 52 & 659 & 122,868 & 99.42 \\
        LLaVA-CoT & 69,006 & 168 & 0 & 68,838 & 99.76 \\
        MMK12 & 15,544 & 0 & 39 & 15,505 & 99.75 \\
        MMR1 & 1,600,235 & 617 & 75,585 & 1,524,033 & 95.24 \\
        PuzzleQA & 1,991 & 0 & 25 & 1,966 & 98.74 \\
        ViRL39K & 36,242 & 37 & 171 & 36,034 & 99.43 \\
        VisualSphinx & 3,776 & 25 & 235 & 3,516 & 93.11 \\
        WaltonColdStart & 51,184 & 17 & 1,381 & 49,786 & 97.27 \\
        WeMath2-Pro & 4,531 & 0 & 197 & 4,334 & 95.65 \\
        WeMath2-SFT & 826 & 0 & 12 & 814 & 98.55 \\
        WeMath2-Standard & 5,683 & 2 & 68 & 5,613 & 98.77 \\
        Zebra-CoT-Physics & 7,035 & 0 & 425 & 6,610 & 93.96 \\
        mmopenr1-8k & 7,428 & 6 & 365 & 7,057 & 95.01 \\
        \midrule
        \textbf{Total} & \textbf{2,372,320} & \textbf{1,806} & \textbf{84,484} & \textbf{2,286,030} & \textbf{96.36} \\
        \bottomrule
    \end{tabular}
    }
    \footnotesize
    \vspace{0.2cm}
    \label{app:datacleaning}
\end{table}

\begin{table}[htbp]
    \centering
    \caption{\textbf{Pass Rate and Consistency Statistics by Dataset.} ``PR'' stands for Pass Rate. The last column shows the count of consistent samples.}
    \label{tab:pass_rate_stats}
    \resizebox{\textwidth}{!}{%
    \begin{tabular}{lrrrc}
        \toprule
        \textbf{Dataset} & \textbf{Rows} & \textbf{PR (Mean)} & \textbf{PR (Med)} & \textbf{Consistent} \\
        \midrule
        BMMR & 80,366 & 0.5272 & 0.5000 & 56,543 \\
        Euclid30K & 26,690 & 0.6976 & 1.0000 & 23,218 \\
        FineVision-ai2d\_merged & 12,167 & 0.8997 & 1.0000 & 10,746 \\
        FineVision-geo170k(qa) & 11,771 & 0.6277 & 0.7500 & 9,130 \\
        FineVision-geometry3k(mathv360k) & 8,977 & 0.4369 & 0.2500 & 4,855 \\
        FineVision-raven & 20,271 & 0.2379 & 0.2500 & 7,994 \\
        FineVision-scienceqa & 6,095 & 0.9028 & 1.0000 & 5,862 \\
        FineVision-tqa & 12,263 & 0.8285 & 1.0000 & 10,568 \\
        FineVision-visualwebinstruct(filt) & 260,556 & 0.5026 & 0.5000 & 162,729 \\
        GameQA-140K & 122,868 & 0.4042 & 0.2500 & 75,002 \\
        LLaVA-CoT & 68,838 & 0.5747 & 0.7500 & 39,516 \\
        MMK12 & 15,505 & 0.7725 & 1.0000 & 14,114 \\
        MMR1 & 1,524,033 & 0.7697 & 1.0000 & 1,293,269 \\
        PuzzleQA & 1,966 & 0.5137 & 0.5000 & 1,445 \\
        ViRL39K & 36,034 & 0.8032 & 1.0000 & 32,651 \\
        VisualSphinx & 3,516 & 0.1778 & 0.0000 & 1,255 \\
        WaltonColdStart & 49,786 & 0.7643 & 1.0000 & 43,233 \\
        WeMath2-Pro & 4,334 & 0.6751 & 1.0000 & 3,368 \\
        WeMath2-SFT & 814 & 0.6225 & 0.7500 & 670 \\
        WeMath2-Standard & 5,613 & 0.7398 & 1.0000 & 4,576 \\
        Zebra-CoT-Physics & 6,610 & 0.7906 & 1.0000 & 6,036 \\
        mmopenr1-8k & 7,057 & 0.5356 & 0.5000 & 4,107 \\
        \midrule
        \textbf{Total} & \textbf{2,286,130} & \textbf{0.6973} & \textbf{1.0000} & \textbf{1,810,887} \\
        \bottomrule
    \end{tabular}
    }
    \footnotesize
    \vspace{0.2cm}
\end{table}
To ensure the high quality and robustness of our training data, we implemented a multi-stage data processing pipeline, consisting of basic structural cleaning and advanced quality assessment. The statistics for these processes are summarized in Table~\ref{tab:data_cleaning_stats} and Table~\ref{tab:pass_rate_stats}.\paragraph{Basic Data Cleaning.}As shown in Table~\ref{tab:data_cleaning_stats}, we first applied rigorous filtering based on sequence length and template validity.The \textbf{"Filt. (Len)"} step removed samples that exceeded the context window limits or were anomalously short, while the \textbf{"Filt. (Tem)"} step discarded samples with parsing errors or malformed templates.Overall, the raw datasets exhibited high structural integrity, with retention rates exceeding 95\% for most subsets. notably, while large-scale datasets like \textit{MMR1} contained a higher absolute number of template errors (75,585), the relative loss remained minimal ($<5\%$).Some specific domains, such as \textit{VisualSphinx} and \textit{Geometry3k}, showed slightly lower retention rates ($\approx 92-93\%$), primarily due to complex formatting requirements inherent to their tasks.
\paragraph{Consistency and Difficulty Analysis.}
Following structural cleaning, we evaluated the semantic quality of the remaining data using a "Pass Rate" (PR) metric and a consistency check, as detailed in Table~\ref{tab:pass_rate_stats}. The Pass Rate serves as an indicator of sample clarity or model competence, where a higher rate implies that the model successfully processed the sample. We observed a clear positive correlation between the Pass Rate and the Consistency Rate:
\begin{itemize}
    \item \textbf{High-Consistency Datasets:} Datasets such as \textit{ScienceQA} and the massive \textit{MMR1} demonstrated exceptional quality, with consistency rates of 96.18\% and 97.30\% respectively, and correspondingly high mean Pass Rates ($\approx 0.8 - 0.9$). This suggests these samples are well-posed and have clear ground truths.
    \item \textbf{Challenging Scenarios:} In contrast, abstract reasoning tasks like \textit{Raven} and \textit{VisualSphinx} exhibited low mean Pass Rates ($<0.25$) and low consistency ($<40\%$). This disparity highlights the inherent difficulty of these tasks or the presence of ambiguous samples that necessitate more robust filtering strategies.
\end{itemize}
Ultimately, we identified approximately 1.81 million consistent samples (Total Consistent) out of the processed pool, providing a solid foundation for stable model training.

\section{Experimental Details}
\subsection{Training Details}

\begin{table}[htbp]
    \centering

    \begin{minipage}[t]{0.32\textwidth}
        \centering
        \caption{SFT Params.}
        \label{tab:sft_params}
        \resizebox{\linewidth}{!}{%
            \begin{tabular}{lc}
                \toprule
                \textbf{Parameter} & \textbf{Value} \\
                \midrule
                Optimizer    & AdamW   \\
                Learning Rate& 1e-5    \\
                Scheduler    & cosine  \\
                Weight Decay & 0.0     \\
                Epochs       & 3       \\
                Warmup Ratio & 3\%     \\
                Sequence Len  & 32,768  \\
                Batch Size   & 32      \\
                Packing      & True    \\
                Liger Kernel & True    \\
                Max Pixels   & $768 \times 768$ \\
                Min Pixels   & $32 \times 32$     \\
                \bottomrule
            \end{tabular}%
        }
    \end{minipage}
    \hfill 
    \begin{minipage}[t]{0.32\textwidth}
        \centering
        \caption{RL Params.}
        \label{tab:rl_params}
        \resizebox{\linewidth}{!}{%
            \begin{tabular}{lc}
                \toprule
                \textbf{Parameter} & \textbf{Value} \\
                \midrule
                Optimizer     & AdamW    \\
                Learning Rate & 1e-6     \\
                Scheduler     & constant \\
                Weight Decay  & 0.1      \\
                Train Steps   & 300     \\
                Warmup Steps  & 10       \\
                Batch Size    & 256      \\
                Prompt Len    & 8192     \\
                Output Len    & 16384    \\
                Temperature   & 1.0      \\
                Rollouts      & 16       \\
                $\epsilon_{\text{low}}, \epsilon_{\text{high}}$ & 3e-4, 4e-4 \\
                \bottomrule
            \end{tabular}%
        }
    \end{minipage}
    \hfill 
    \begin{minipage}[t]{0.32\textwidth}
        \centering
        \caption{Eval Params.}
        \label{tab:eval_params}
        \resizebox{\linewidth}{!}{%
            \begin{tabular}{lc}
                \toprule
                \textbf{Parameter} & \textbf{Value} \\
                \midrule
                Engine       & VLLM     \\
                Precision    & BF16     \\
                Temperature  & 0.0      \\
                Top-p        & 1.0      \\
                Top-k        & -1       \\
                Max Tokens   & 32768     \\
                Rep. Penalty & 1.05      \\
                Max Pixels   & $2048 \times 2048$ \\
                Min Pixels   & $32 \times 32$     \\
                \bottomrule
            \end{tabular}%
        }
    \end{minipage}
\end{table}
We provide a comprehensive overview of our experimental setup, including the hyperparameters for Supervised Fine-Tuning (SFT), Reinforcement Learning (RL).

\paragraph{Supervised Fine-Tuning (SFT).}
As shown in Table~\ref{tab:sft_params}, the SFT stage utilizes a cosine learning rate scheduler with a peak learning rate of $1\text{e-}5$. To optimize training efficiency and memory usage, we employ sequence packing with a length of 32,768 and integrate Liger Kernel support. We also accommodate high-resolution inputs by setting the maximum pixel limit to $768 \times 768$.

\paragraph{Reinforcement Learning (RL).}
Following SFT, we further align the model using Reinforcement Learning. The detailed hyperparameters are listed in Table~\ref{tab:rl_params}. In this stage, we adopt a constant learning rate of $1\text{e-}6$ to ensure stability. The training process involves 16 rollouts per prompt with a KL-divergence penalty controlled by $\epsilon_{\text{low}}=3\text{e-}4$ and $\epsilon_{\text{high}}=4\text{e-}4$. The input and output max lengths are set to 8,192 and 16,384 tokens, respectively.

\subsection{Evaluation Details}
\label{apx:eval}
We conduct a comprehensive evaluation using the VLMEvalKit~\citep{vlmevalkit}.  For evaluation metrics, we replace traditional exact string matching with the compass-verifier~\citep{liu2025compassverifierunifiedrobustverifier}, which employs an LLM-as-a-Judge to accurately assess response correctness. Regarding rollout settings, following the guidelines from OpenDataArena~\citep{cai2025opendataarenafairopenarena} and the official Qwen documentation, we set the temperature to $0.0$, top-p to $1.0$, and top-k to $-1$. We apply a repetition penalty of $1.05$ and limit the maximum response length to $32768$ tokens. Furthermore, we exclude all system prompts during inference.

\clearpage
\section{Prompts}
\label{sec:prompts}
In this section, we provide the full prompts used in our experiments.

\subsection{Question Cleaning}
\begin{tcolorbox}[
  colback=gray!5,
  colframe=black!75,
  width=\textwidth,
  title=Question Cleaning Prompt,
  breakable
]
\small

\textbf{Task.} Clean the given question text by following these steps.

\medskip
\textbf{Error types}

\textbf{1. Translation}
\begin{itemize}
  \item Translate the question into English.
  \item If the question is already in English, keep it as is.
\end{itemize}

\textbf{2. Irrelevant content}
\begin{itemize}
  \item Remove all irrelevant links, advertisements, signatures, emails, special symbols, or repeated punctuation.
  \item Remove Markdown watermarks, unrelated tables, or redundant markings in formulas.
  \item Remove question numbers, IDs, or scoring information that appears before the actual question (e.g., ``Q12.'', ``(5 points)'', ``1.'').
  \item Do \emph{not} consider an \texttt{<image>} tag at the beginning of a question as irrelevant content.
\end{itemize}

\textbf{3. Non-answer question}
\begin{itemize}
  \item Questions that are not actual problem-solving questions, e.g., asking to draw a diagram or write code, should be marked under this error type.
  \item Questions that are incomplete to the point that they cannot be answered.
\end{itemize}

\textbf{4. Low-quality instruction}
\begin{itemize}
  \item If the question contains instructions that may reduce reasoning quality (e.g., ``just give the answer'', ``do not think'', ``give me the final answer only''), rewrite these into instructions that encourage thoughtful reasoning (e.g., ``provide a clear reasoning process before the final answer'').
\end{itemize}

\medskip
\textbf{Output rules}

\begin{itemize}
  \item If the question has no issues (no translation or cleaning needed), output:
  \begin{quote}
    \texttt{No Problem}
  \end{quote}

  \item If the question has issues, output JSON in the following format:
  \begin{quote}
    \texttt{\{}\\
    \texttt{"error\_type": ["translation", "irrelevant content"], // only "translation", "irrelevant content", "non-answer question" and "low-quality instruction"}\\
    \texttt{"corrected\_text": "Cleaned and translated version of the question (leave empty if non-answer question)"}\\
    \texttt{\}}
  \end{quote}
\end{itemize}

\medskip
\textbf{Examples}

\textbf{Example 1}

\textit{Input:}
\begin{quote}
What is this? https://example.com
\end{quote}

\textit{Output:}
\begin{quote}
\texttt{\{}\\
\texttt{"error\_type": ["translation", "irrelevant content"],}\\
\texttt{"corrected\_text": "What is this?"}\\
\texttt{\}}
\end{quote}

\medskip
\textbf{Example 2}

\textit{Input:}
\begin{quote}
Please directly answer the question: what are the roots of this equation?
\end{quote}

\textit{Output:}
\begin{quote}
\texttt{\{}\\
\texttt{"error\_type": ["translation", "low-quality instruction"],}\\
\texttt{"corrected\_text": "Please provide a clear reasoning process before giving the roots of this equation."}\\
\texttt{\}}
\end{quote}

\medskip
\textbf{Example 3}

\textit{Input:}
\begin{quote}
\{question\}
\end{quote}

\textit{Output:}
\begin{quote}
\texttt{No Problem}
\end{quote}

\end{tcolorbox}

\captionof{table}{Prompt for question cleaning.}
\label{pmt:clean}

\subsection{Answer Extraction}
\begin{tcolorbox}[
  colback=gray!5,
  colframe=black!75,
  width=\textwidth,
  title=Answer Extraction Prompt,
  breakable
]
\small

You are a precise math answer extractor. Your task is to read the user’s question and the provided solution, then extract ONLY the final answer(s).

Output EXACTLY one \texttt{<answer>...</answer>} tag containing only the final answer, with no extra text or explanations.

\medskip
\textbf{Extraction Rules (Follow in order of priority)}

\begin{enumerate}
  \item \textbf{Top Priority (\(\boxed{\cdots}\)).} If a final \(\boxed{\cdots}\) is present, output its INNER CONTENT EXACTLY as written, preserving all LaTeX, symbols, and text. This rule takes precedence over all other rules (including the unit rule). Ensure the extracted content is complete (e.g., balanced braces).

  \item \textbf{Final Result (No Box).} If no \(\boxed{\cdots}\) is found, extract the final explicit numerical or symbolic result (e.g., after ``final answer is'', ``answer is'', ``Thus'', ``Therefore'').

  \item \textbf{LaTeX Preservation.} When applying Rule 2, preserve all LaTeX expressions and symbols (e.g., \verb|\sqrt{...}|, \(\infty\), \(\frac{\cdots}{\cdots}\), \(\pi\)). Do NOT convert LaTeX to plain numbers or words.

  \item \textbf{No Simplification.} Do NOT convert words to digits, rewrite mixed numbers, or simplify fractions unless they already appear that way in the final result.

  \item \textbf{Unit Stripping (No Box Only).} If applying Rule 2 (i.e., no \(\boxed{\cdots}\) was found), do NOT include units (e.g., cm, dollars, ways). Exception: Always keep the percent sign (\%).

  \item \textbf{Multiple Solutions.} If the final answer lists multiple distinct values (e.g., ``\(x=5\) or \(x=10\)'', ``the roots are \(-1\) and \(1\)''), output them as a single, comma-separated string (e.g., \texttt{"5, 10"}, \texttt{"-1, 1"}).

  \item \textbf{Word Answers.} If the solution's final answer is a definitive word (e.g., ``Yes'', ``No'', ``True'', ``False'', ``None'', ``Cannot be determined''), extract that word.

  \item \textbf{Not Found.} If no specific, concise answer (mathematical, expression, or definitive word) can be found, respond with \texttt{<answer></answer>}.
\end{enumerate}

\medskip
\textbf{Examples}

\textbf{Example 1 (Boxed).}

Solution: \texttt{...blah blah... The answer is $\boxed{10}$.} \\
Respond with: \texttt{<answer>10</answer>}

\medskip
\textbf{Example 2 (Boxed with LaTeX).}

Solution: \texttt{...so the value is $\boxed{\frac{\sqrt{3}}{2}}$.} \\
Respond with: \texttt{<answer>\textbackslash frac\{\textbackslash sqrt\{3\}\}\{2\}</answer>}

\medskip
\textbf{Example 3 (Boxed with Units --- Rule 1 Precedence).}

Solution: \texttt{...the final area is $\boxed{24 \text{ cm}^2}$.} \\
Respond with: \texttt{$<answer>24 \text\{ cm\}^2</answer>$}

\medskip
\textbf{Example 4 (No Box with Units --- Rule 5 Applies).}

Solution: \texttt{...Therefore, the length is 40 meters.} \\
Respond with: \texttt{<answer>40</answer>}

\medskip
\textbf{Example 5 (No Box with Percent --- Rule 5 Exception).}

Solution: \texttt{...The total increase was 15.5\%.} \\
Respond with: \texttt{<answer>15.5\%</answer>}

\medskip
\textbf{Example 6 (Multiple Solutions).}

Solution: \texttt{...the roots of the equation are $x = -2$ or $x = 5$.} \\
Respond with: \texttt{<answer>-2, 5</answer>}

\medskip
\textbf{Example 7 (Word Answer).}

Solution: \texttt{...we can conclude that the statement is False.} \\
Respond with: \texttt{<answer>False</answer>}

\medskip
\textbf{Example 8 (Not Found).}

Solution: \texttt{...this completes the proof by induction.} \\
Respond with: \texttt{<answer></answer>}

\medskip
\textbf{Task template}

\medskip
Question: \texttt{\{instruction\}} \\[0.3em]
Solution: \texttt{\{output\_tail\}} \\[0.3em]
Respond with: \texttt{<answer>...</answer>}

\end{tcolorbox}

\captionof{table}{Prompt for Answer Extraction.}
\label{pmt:answer_extraction}

\subsection{Image Captioning}
\begin{tcolorbox}[
  colback=gray!5,
  colframe=black!75,
  width=\textwidth,
  title=Multimodal Data Annotation Specialist,
  breakable
]
\small

You are a meticulous \textbf{Multimodal Data Annotation Specialist}. Your primary mission is to deconstruct multimodal tasks (consisting of images and text) and translate them into a highly structured and comprehensive natural language description. The goal is to create a ``golden'' reference text that is as unambiguous and detailed as a data file, which will be used to evaluate the accuracy of other AI models. Your adherence to the format described below is critical. You will be provided with a task that consists of up to two parts: one \textbf{image}, and its corresponding \textbf{question text}.

\medskip
\textbf{Guiding Principles for Analysis:}

\begin{enumerate}
  \item \textbf{Category-First, Structure-Always:}  
  Your entire analysis begins with correctly identifying the image's category. The category list now includes \textbf{both STEM-style images and natural photographs} (see expanded list below). This category dictates the focus of your description. You must then follow the specified markdown structure precisely for your output.

  \item \textbf{Separate What is Seen from What is Inferred:}  
  Your description must maintain a strict separation between elements explicitly visible in the image and properties inferred from the accompanying text. The output format has dedicated sections for this.

  \item \textbf{Comprehensive and Atomic Breakdown:}  
  Every single element in the image must be described individually within the ``Explicit Component Breakdown'' section. For natural images, this includes:
  \begin{itemize}
    \item People (pose, clothing, objects held)
    \item Everyday objects
    \item Scene elements (furniture, roads, sky, vehicles)
    \item Background structures
    \item Animals, plants, food, tools, etc.  
          Treat each as a standalone component.
  \end{itemize}

  \item \textbf{Holistic Synthesis:}  
  The image and question text are a single unit. Use the text to define roles, identify actions, or extract inferred properties.
\end{enumerate}

\medskip
\textbf{Instructions for Structuring Your Output}

You must generate a single text block. The response must be structured using markdown with headings, \textbf{bolded keywords}, and bullet points \textbf{exactly as specified below}.

For each image provided, create a complete descriptive block starting with:

\medskip
\textbf{\#\#\# Image [N]: [Primary Category Name]}

\medskip
\textbf{Required Output Structure:}

\begin{itemize}
  \item \textbf{Heading.} \texttt{\#\#\# Image [N]: [Primary Category Name]} (replace \texttt{[N]} with the image number, and \texttt{[Primary Category Name]} with the category you identify from the list below).

  \item \textbf{Scene Summary.}  
  A single, concise sentence that describes the overall purpose and content of the image.

  \item \textbf{Explicit Component Breakdown.}  
  (This section is for \textbf{visible elements only}.)
  \begin{itemize}
    \item \texttt{[Component Name] (`[label]`):} A description of the component. The \texttt{[label]} should be the exact text or symbol labeling the component in the image. If there is no label, use \texttt{None}.
    \item Repeat for every single visible component: objects, vectors, surfaces, axes, points, everyday objects, people, clothing, background structures, etc.
  \end{itemize}

  \item \textbf{Interactions and Relationships.}  
  (This section describes how the explicit components are connected and arranged.)
  \begin{itemize}
    \item Describe spatial and structural connections (e.g., ``Person A stands next to table B'', ``Button `A' triggers Modal `B'''').
    \item Describe logical or physical relationships (e.g., contact, holding, containment, occlusion).
    \item Trace directional flows (arrows/process steps) or describe data trends (charts/graphs).
  \end{itemize}

  \item \textbf{Implicit and Inferred Properties.}  
  (This section is \textbf{only} for information derived from the question text or domain conventions, not explicitly drawn.)
  \begin{itemize}
    \item \textbf{[Component or System Name]:} [Inferred Property]. For example, \textbf{Person A:} identified as ``teacher'' from question text.
    \item \textbf{[Component or System Name]:} [Inferred Property]. For example, \textbf{Dataset:} values normalized to \texttt{[0, 1]}.
    \item List every piece of non-visual information here.
  \end{itemize}

  \item \textbf{Identified Ambiguities.}  
  (If any part of the image is illegible or unclear, list it here. If none, state ``None''.)
  \begin{itemize}
    \item \texttt{[Description of ambiguous element]}.
  \end{itemize}
\end{itemize}

\medskip
\textbf{Reference Guide: Image Categories}

Below is the expanded and unified category list.

\medskip
\textbf{STEM / Diagrammatic Categories}

\begin{itemize}
  \item Geometric Diagram
  \item Spatial Reasoning Scene
  \item Mathematical Plot / Chart
  \item Puzzle / Logic Diagram
  \item Textbook Illustration
  \item Physics / Mechanics Diagram
  \item Experimental Setup
  \item Astronomy / Space Visualization
  \item Molecular / Chemical Diagram
  \item Biological Structure
  \item Geological / Earth Science Diagram
  \item Circuit / Network Diagram
  \item Abstract Mathematical Representation
  \item Table / Matrix
  \item Diagram / Flowchart
\end{itemize}

\medskip
\textbf{Natural Image Categories}

\begin{itemize}
  \item Natural Landscape Scene
  \item Urban / Street Scene
  \item Indoor / Interior Scene
  \item Human Portrait / Activity
  \item Sports / High-Motion Scene
  \item Animal / Wildlife Scene
  \item Product / Still Life Object
  \item Vehicle / Machinery Object
  \item Food / Beverage Item
  \item Document / Text Image
  \item Artwork / Illustration
  \item Technical / Surveillance / Medical
\end{itemize}

\medskip
\textbf{Now, analyze the provided image(s) and question text}, and produce the structured natural language description in this exact format:

\medskip
\texttt{\{question\}}\\
\texttt{\{image\}}

\end{tcolorbox}
\captionof{table}{Prompt for image captioning.}
\label{pmt:caption}

\subsection{Long CoT Distillation}
\begin{tcolorbox}[
  colback=gray!5,
  colframe=black!75,
  width=\textwidth,
  title=Long CoT Distillation Prompt,
  breakable
]
\small

You are an expert in science and visual reasoning with advanced capabilities in multimodal analysis.
Your response will be used as a high-quality example to train a new AI model.
Solve the problem efficiently and clearly by integrating \textbf{all} information from multimodal inputs.

\medskip
\textbf{Core Principles}

\begin{enumerate}
  \item \textbf{Equal Weight to All Inputs.}
  Information from images (photos, charts, graphs, diagrams, tables, handwritten notes) is as important as text. Never ignore visual elements.

  \item \textbf{Systematic Analysis.}
  Follow a rigorous, reproducible approach for every problem.

  \item \textbf{Precision and Accuracy.}
  Double-check all calculations and reasoning steps.

  \item \textbf{Adaptive Reasoning.}
  Choose the most appropriate method based on the specific problem context.
\end{enumerate}

\medskip
\textbf{Solution Framework}

\medskip
\textbf{Phase 1: Comprehensive Information Extraction}

\begin{itemize}
  \item Carefully analyze all text content for requirements, constraints, and given values.
  \item Thoroughly examine all visual elements, extracting every piece of relevant information.
  \item Note measurements, relationships, patterns, and any implicit information.
  \item Explicitly connect visual and textual information when they relate to each other.
\end{itemize}

\medskip
\textbf{Phase 2: Strategic Problem Setup}

\begin{itemize}
  \item Compile all extracted information in an organized manner.
  \item Clearly state what needs to be found or proven.
  \item Identify the most relevant scientific principles and methodologies.
  \item Consider what assumptions may be necessary and state them explicitly.
\end{itemize}

\medskip
\textbf{Phase 3: Rigorous Solution Execution}

\begin{itemize}
  \item Present your solution with complete logical flow.
  \item Show all mathematical steps with proper notation.
  \item When using formulas, present them clearly, substitute values, and then calculate.
  \item Reference specific parts of visual inputs when they support your reasoning.
  \item Maintain unit consistency throughout all calculations.
  \item Keep appropriate precision and significant figures.
\end{itemize}

\medskip
\textbf{Phase 4: Solution Validation}

\begin{itemize}
  \item Verify your answer makes scientific and logical sense.
  \item Check that all parts of the question have been addressed.
  \item For multiple choice questions, confirm your selection and briefly justify if needed.
  \item Ensure dimensional analysis is correct.
\end{itemize}

\medskip
\textbf{Key Reminders}

\begin{itemize}
  \item Visual information is never supplementary --- it is integral to the solution.
  \item Every piece of data from images must be considered.
  \item Your reasoning should be so clear that someone could follow it without seeing the images.
  \item When in doubt, show more work rather than less.
  \item Connect each step logically to build a complete solution narrative.
\end{itemize}

\medskip
\textbf{Answer Format Guidelines}

Determine the nature of your answer:

\begin{itemize}
  \item \textbf{If the problem has a definitive, fixed answer} (numerical value, specific choice, exact result):
  \begin{itemize}
    \item Present your complete reasoning and solution process.
    \item At the end, clearly state: \\
          \texttt{Therefore, the final answer is <answer>YOUR\_ANSWER</answer>}
          (with the actual answer substituted for \texttt{YOUR\_ANSWER}).
    \item Examples: \texttt{<answer>5.2 m/s</answer>}, \texttt{<answer>C</answer>}, \texttt{<answer>2.5 m, 30°</answer>}.
  \end{itemize}

  \item \textbf{If the problem requires explanation, discussion, or has no single fixed answer}:
  \begin{itemize}
    \item Focus on presenting your points clearly and in a structured manner.
    \item Provide a full analysis and explanation.
    \item You may include examples, reasoning steps, or possible conclusions, but a single ``correct'' answer wrapped with \texttt{<answer>} tags is not required.
  \end{itemize}
\end{itemize}

\medskip
\textbf{Problem injection}

The problem will be inserted as:

\medskip
\texttt{Problem: \{item['question']\}}

\medskip
Analyze all provided materials carefully, think through the problem step by step, and provide a comprehensive solution that demonstrates mastery of both scientific reasoning and visual analysis.

\medskip
\textbf{Final line constraint}

The last line of your response must be exactly:

\medskip
\texttt{"Therefore, the final answer is <answer>ANSWER</answer>."}

\end{tcolorbox}

\captionof{table}{Prompt for long chain-of-thought distillation.}
\label{pmt:distill}

\subsection{Answer Verification}
\begin{tcolorbox}[
  colback=gray!5,
  colframe=black!75,
  width=\textwidth,
  title=Answer Verification Prompt,
  breakable
]
\small

You are an expert evaluator. Compare the reference and generated answers only for semantic correctness and factual agreement.

\medskip
\textbf{Task}  
Determine whether the two answers express the same correct solution. Focus on meaning, correctness, and final results rather than wording or format.

\medskip
\textbf{Evaluation Guidences}
\begin{itemize}
  \item \textbf{Equivalent}: same conclusion or final answer, no substantive factual differences.
  \item \textbf{Different}: conflicting conclusions, missing required reasoning, or any factual mistake in the generated answer.
\end{itemize}

\medskip
\textbf{Input Question}

\texttt{\{question\}}

\medskip
\textbf{Reference Answer}

\texttt{\{solution\}}

\medskip
\textbf{Generated Answer}

\texttt{\{response\}}

\medskip
\textbf{Output Intructions}  
\\Respond in the following two-line format (no extra text):
\begin{quote}
Analysis:  \texttt{\textless}concise reasoning\texttt{\textgreater} \\
Judgment: \texttt{\textless}Equivalent or Different \texttt{\textgreater}
\end{quote}

\end{tcolorbox}

\captionof{table}{Prompt for Answer Verification}
\label{pmt:verify}

\section{Filteration Cases}
\label{apx:case}
In this section, we provide representative examples for each error type used in our question cleaning stage, including \emph{translation}, \emph{irrelevant content}, \emph{non-answer question}, and \emph{low-quality instruction}. These cases are drawn from multiple sources such as BMMR, Euclid30K, and VisualSphinx.

\subsection{Translation}

\begin{tcolorbox}[
  colback=gray!5,
  colframe=black!75,
  width=\textwidth,
  title=Case: Translation (Euclid30k-16648),
  breakable
]
\small

\begin{center}
\includegraphics[width=0.4\linewidth]{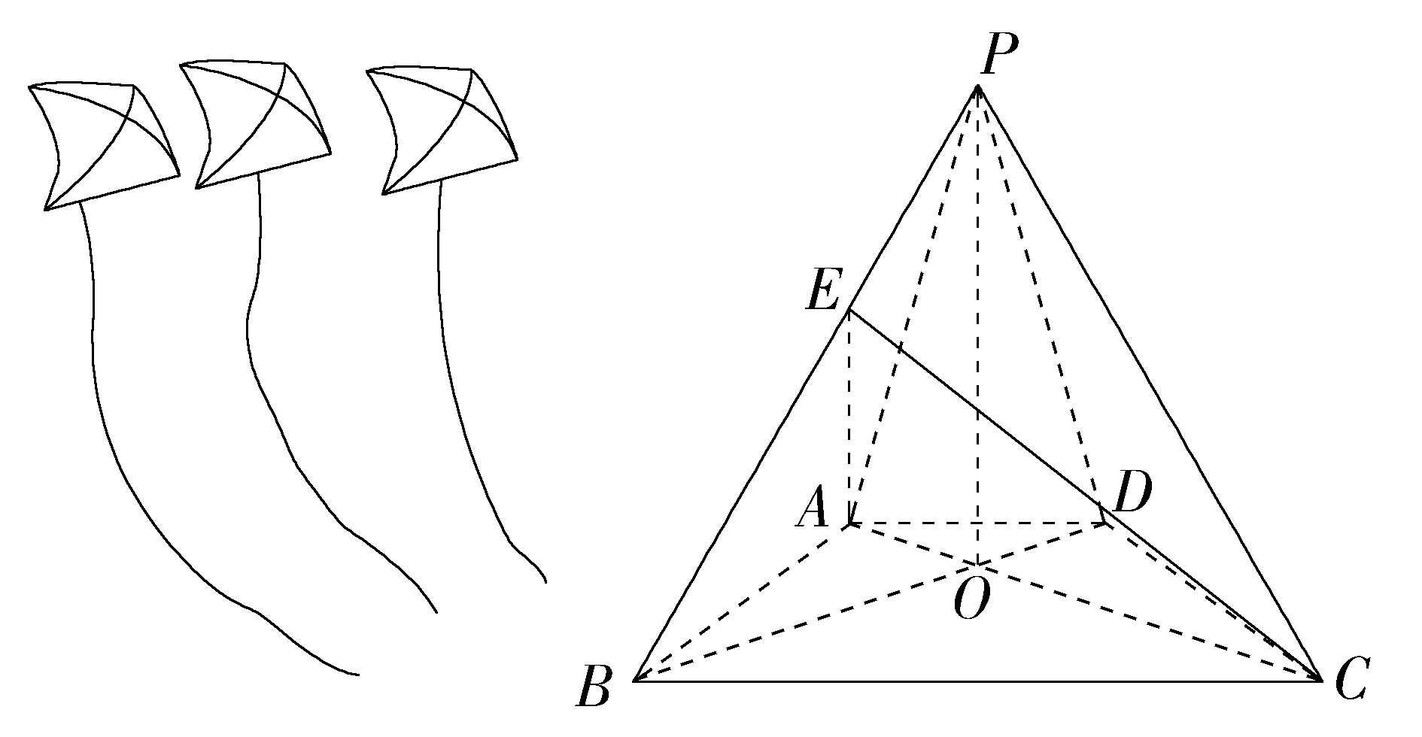}
\end{center}

\medskip

\textbf{Question:}

\medskip
\begin{CJK*}{UTF8}{gbsn} 
如图，某种风筝的骨架模型是四棱锥P-ABCD，四边形ABCD是等腰梯形，AD \(\parallel\) BC，AC \(\cap\) DB=O，PO \(\perp\) 平面ABCD，\(\angle\) BOC=$90^\circ$，\end{CJK*}\begin{CJK*}{UTF8}{gbsn} OA=1，OC=2，E在PB上。\\
(2)研究表明，当\(\tan \angle PAC=2\)时，风筝表现最好，求此时直线PA与平面PBC所成角的正弦值。
\end{CJK*}
\medskip

\textbf{Error Type:}
Translation

\textbf{Corrected Question:}

\medskip

The frame model of a certain kite is a quadrilateral pyramid P-ABCD, where quadrilateral ABCD is an isosceles trapezoid with AD parallel to BC. The diagonals AC and BD intersect at point O, and PO is perpendicular to the plane ABCD. It is given that angle BOC = 90°, OA = 1, and OC = 2. Point E lies on PB.

(2) Research shows that the kite performs best when tan(angle PAC) = 2. Find the sine of the angle between line PA and plane PBC at that time.

\medskip

\textbf{Filtering Reason:}

\medskip

The original question is in Chinese and contains a specific reference to the diagram. The translation was needed to ensure the question was clearly understandable in English. The format was adjusted to correct the translation and ensure clarity, maintaining the integrity of the mathematical context.

\end{tcolorbox}

\captionof{table}{Case of translation (Euclid30k-16648).}

\subsection{Irrelevant content}
\begin{tcolorbox}[
  colback=gray!5,
  colframe=black!75,
  width=\textwidth,
  title=Case: Irrelevant content (BMMR-84693),
  breakable
]
\small

\begin{center}
\includegraphics[width=0.4\linewidth]{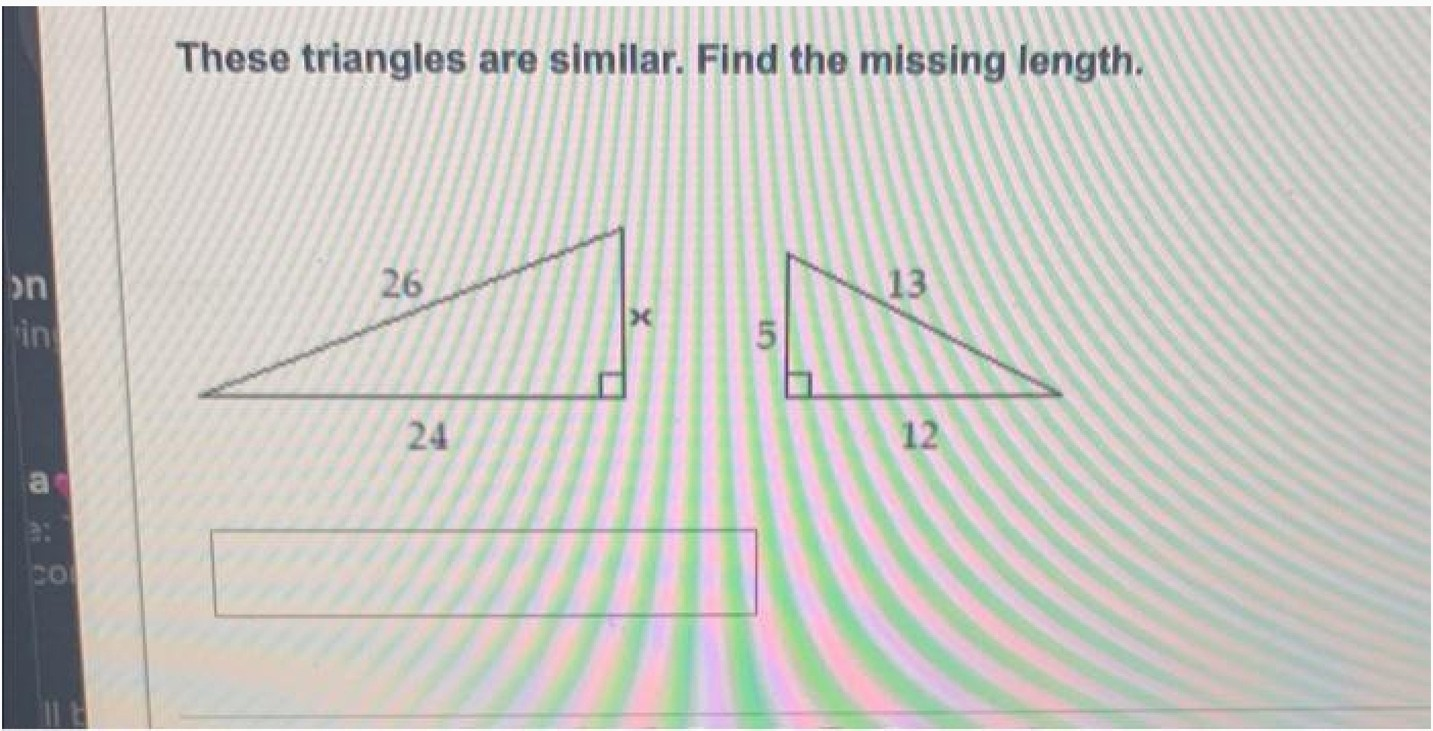}
\end{center}

\medskip

\textbf{Question:}

\medskip

\texttt{<image>} These triangles are similar. Find the missing length.\\[0.5em]
\texttt{![Two similar right - angled triangles. The first has hypotenuse 26, base 24 and height \$x\$. The second has hypotenuse 13, base 12 and height 5.](}\url{https://latex.artofproblemsolving.com/tex2html/69101954001041251272727742444710200451977070775401974217746220747051752954015457019440...}\texttt{)}

\medskip

\textbf{Error Type:}
Irrelevant content 

\medskip

\textbf{Corrected Question:}

\medskip

These triangles are similar. Find the missing length.

\medskip

\textbf{Filtering Reason:}

\medskip

The original question contains a long Markdown image caption and an external URL that redundantly describe the diagram in text form. This information is not needed to understand or solve the problem, since the visual content is already provided by the image itself. We therefore remove this caption--URL block as \emph{irrelevant content} and keep only the core question text, without changing the mathematical meaning of the task.

\end{tcolorbox}

\captionof{table}{Case of irrelevant content (BMMR-84693).}

\subsection{Non-answer question}
\begin{tcolorbox}[
  colback=gray!5,
  colframe=black!75,
  width=\textwidth,
  title=Case: Non-answer question (BMMR-72354),
  breakable
]
\small

\begin{center}
\includegraphics[width=0.35\linewidth]{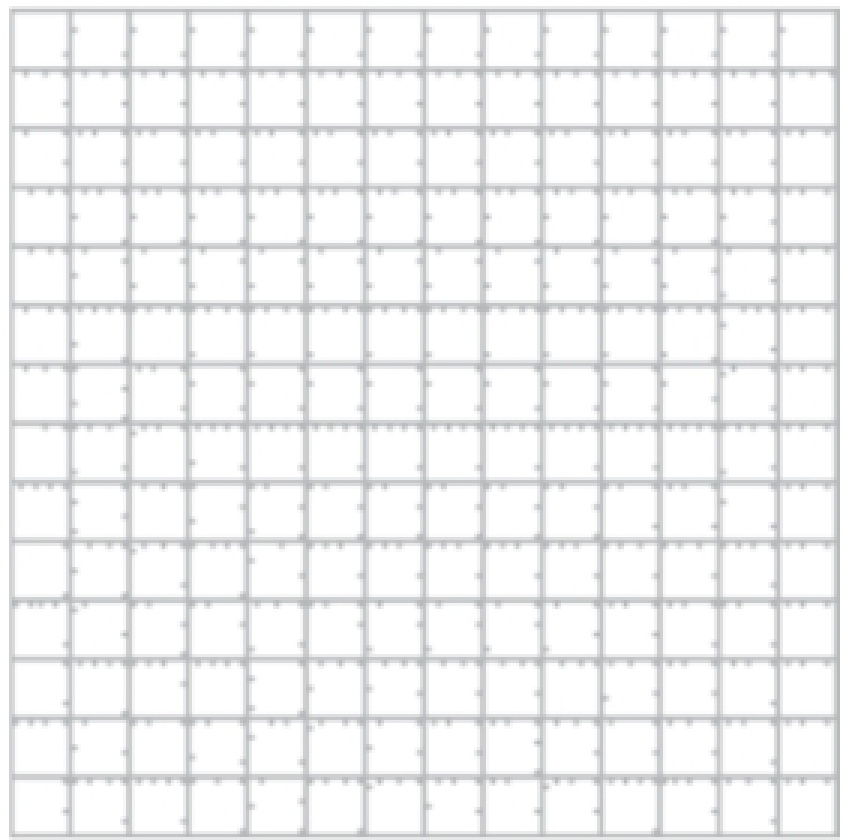}
\end{center}

\medskip

\textbf{Question:}

\texttt{<image>} 1. A rectangle has a perimeter of 444 centimeters and a side of length l centimeters.  
   - a. Draw a rectangle to represent this situation. Label each side of the rectangle in terms of $l$.  
   - b. Write an equation for the area A of the rectangle in terms of $l$.  
   - c. Sketch a graph of the relationship between the length of a side and the area. Be sure to include important features such as the maximum/minimum, y-intercept(s), x-intercept(s), and the line of symmetry.  
   - d. Study your graph, table, and equation for the area of rectangles with a fixed perimeter of 444 centimeters. Which representation is the most useful for predicting the maximum area? Explain your choice.

\medskip

\textbf{Error Type:}
Non-answer question

\textbf{Response:}

\texttt{\{}\\
\texttt{"error\_type": ["non-answer question"],}\\
\texttt{"corrected\_text": ""}\\
\texttt{\}}

\medskip

\textbf{Filtering Reason:}

This question involves drawing and writing code, which is not an answerable question. It asks the student to graph and sketch diagrams as well as explain reasoning, but it doesn't provide any actual problem to solve or compute. Therefore, this is a non-answer question, and the \texttt{corrected\_text} is left empty.

\end{tcolorbox}

\captionof{table}{Case of non-answer question (BMMR-72354).}

\begin{tcolorbox}[
  colback=gray!5,
  colframe=black!75,
  width=\textwidth,
  title=Case: Non-answer question (VisualSphinx-631),
  breakable
]
\small

\begin{center}
\includegraphics[width=0.35\linewidth]{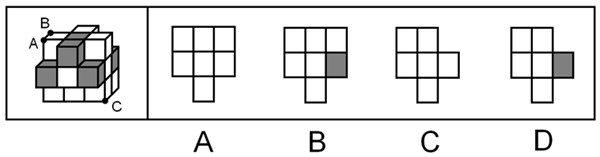}
\end{center}

\medskip

\textbf{Question:}

The following diagram is a polyhedron composed of 15 white and 5 gray cubes, after its [translation incomplete, seems to be cut off] --- Options: A) A \quad B) B \quad C) C \quad D) D

\medskip

\textbf{Error Type:}
Non-answer question

\textbf{Response:}

\texttt{\{}\\
\texttt{"error\_type": ["non-answer question"],}\\
\texttt{"corrected\_text": ""}\\
\texttt{\}}

\medskip

\textbf{Filtering Reason:}

The question text is explicitly marked as ``translation incomplete'' and is cut off after the phrase ``after its ...'', without specifying what operation or condition is applied to the polyhedron. Only the answer options (A, B, C, D) are listed, but the task to be solved is missing. As a result, the problem statement is incomplete and no definite answer can be determined, so this sample is labeled as a \emph{non-answer question} and the \texttt{corrected\_text} field is left empty.

\end{tcolorbox}

\captionof{table}{Case of non-answer question (VisualSphinx-631).}

\subsection{Low-quality instruction}
\begin{tcolorbox}[
  colback=gray!5,
  colframe=black!75,
  width=\textwidth,
  title=Case: Low-quality instruction (LLaVA-CoT-100K-1761),
  breakable
]
\small

\begin{center}
\includegraphics[width=0.4\linewidth]{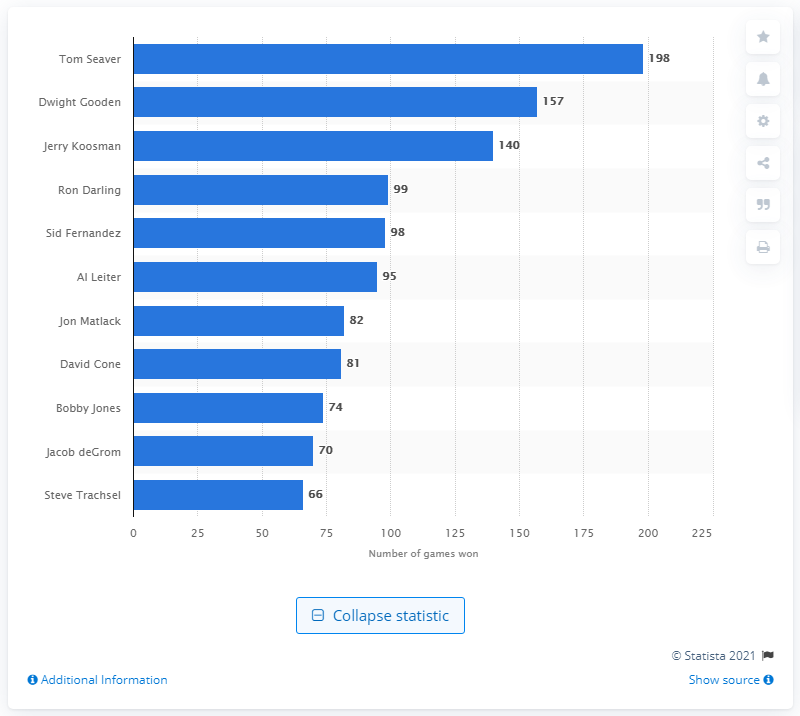}
\end{center}

\medskip

\textbf{Question:}

\medskip

Who has won the most games in New York Mets franchise history? Answer the question using a single word or phrase.

\medskip

\textbf{Error Type:}
Low-quality instruction

\textbf{Corrected Question:}

\medskip

Who has won the most games in New York Mets franchise history? Please provide a clear reasoning process before giving the final answer.

\medskip

\textbf{Filtering Reason:}

\medskip

The original question contains the instruction to answer with a single word or phrase, which could limit the reasoning process. This reduces the quality of the question by potentially discouraging a more comprehensive reasoning approach. To encourage better reasoning and a more thoughtful answer, the corrected question requests a clear reasoning process before providing the final answer.

\end{tcolorbox}

\captionof{table}{Case of low-quality instruction (LLaVA-CoT-100K-1761).}

\end{document}